\def\year{2021}\relax
\documentclass[letterpaper]{article} 
\usepackage{aaai21}  
\usepackage{times}  
\usepackage{helvet} 
\usepackage{courier}  
\usepackage[hyphens]{url}  
\usepackage{graphicx} 
\usepackage{natbib}  
\usepackage{caption} 
\usepackage[utf8]{inputenc} 
\usepackage[T1]{fontenc}    
\usepackage{hyperref}       
\usepackage{url}            
\usepackage{booktabs}       
\usepackage{amsfonts}       
\usepackage{nicefrac}       
\usepackage{microtype}      

\usepackage{mathtools}
\usepackage{amsmath}

\usepackage{tcolorbox}
\newtcolorbox{mybox}{left=2.5pt,right=2.5pt}
\usepackage[ruled,vlined]{algorithm2e}
\usepackage{subcaption}
\newcommand \numberthis{\addtocounter{equation} {1} \tag{\theequation}}

\DeclareMathOperator*{\argmin}{arg\,min}
\newcommand{\ignore}[1]{}
\usepackage[switch]{lineno}
\title{Exacerbating Algorithmic Bias through Fairness Attacks}

\author{%
Ninareh Mehrabi\textsuperscript{\rm 1,2}, Muhammad Naveed\textsuperscript{\rm 1}, Fred Morstatter\textsuperscript{\rm 1,2}, Aram Galstyan\textsuperscript{\rm 1,2} \\
}
\affiliations{
\textsuperscript{\rm 1}University of Southern California - \textsuperscript{\rm 2}Information Sciences Institute \\
\{ninarehm, mnaveed\}@usc.edu, \{fredmors, galstyan\}@isi.edu }
\begin{document}

\maketitle
\begin{abstract}
Algorithmic fairness has attracted significant attention in recent years, with many quantitative measures suggested for characterizing the fairness of different machine learning algorithms. Despite this interest, the robustness of those fairness measures with respect to an intentional adversarial attack has not been properly addressed. Indeed, most adversarial machine learning has focused on the impact of malicious attacks on the accuracy of the system, without any regard to the system's fairness. We propose new types of data poisoning attacks where an adversary intentionally targets the fairness of a system. Specifically, we propose two families of attacks that target fairness measures. In the \textit{anchoring attack}, we skew the decision boundary by placing poisoned points near specific target points to bias the outcome. In the \textit{influence attack on fairness}, we aim to maximize the covariance between the sensitive attributes and the decision outcome and affect the fairness of the model. We conduct extensive experiments that indicate the effectiveness of our proposed attacks.
\end{abstract}

\section{Introduction}

With proliferation of machine learning (ML) applications in everyday life, it is imperative that ML algorithms underlying those applications do not discriminate, especially when it comes to potentially sensitive and consequential decisions, such as bail decisions~\cite{Dresseleaao5580}. Thus,  recent research has looked into possible biases present in ML algorithms, and proposed different measures and definitions  for characterizing fairness \cite{dwork2012fairness,hardt2016equality,NIPS2017_6995,10.1145/3194770.3194776,mehrabi2019survey}. 

Despite this interest, not much is known about the robustness of various fairness measures with respect to random, or perhaps malicious, perturbations. Indeed, it is known that machine learning models can be susceptible to various types of adversarial attacks targeted to degrade the performance of machine learning models. However, research in adversarial machine learning has mostly focused on targeting accuracy \cite{chakraborty2018adversarial,li2018security}. We argue that, like accuracy, fairness measures can be targeted by malicious adversaries as well. For instance, adversaries can attack models used by a government agency with the goal of making them appear unfair in order to depreciate their value and credibility. Some adversaries can even profit from such attacks by biasing decisions for their benefit, e.g., in credit or loan applications. Thus, one should consider fairness when assessing the robustness of ML systems.

\textbf{Our contributions.} In this work, we propose data poisoning attacks that target fairness. We propose two families of poisoning attacks: {\em anchoring} and {\em influence}\footnote{https://github.com/Ninarehm/attack}. In anchoring attacks the goal is to place poisoned points to affect fairness without modifying the attacker loss. On the other hand, our influence attack on fairness can affect both fairness and accuracy by injecting poisoned points during train time via a specific adversarial loss that regularizes between fairness and accuracy losses. Some adversaries may want to harm systems with regard to fairness and accuracy at the same time, while others might only consider one that can be achieved by this regularization. In the anchoring attack, we place poisoned points to bias the decision boundary; in the influence attack, we target fairness measures by incorporating a loss function maximizing and attacking which can degrade fairness by maximizing the covariance between the decision outcome and sensitive attributes.

Through experimentation on three different datasets with different fairness measures and definitions, we show the effectiveness of our attacks in achieving the desired goal of affecting fairness. In addition, we incorporate different baseline models to evaluate different aspects of our attacks. We demonstrate that original data poisoning attacks designed to attack accuracy are not suitable for fairness attacks, thus highlighting the importance of attacks designed for fairness. We also compare our methods against concurrent work on adversarial attacks on fairness and show the effectiveness of our methods in comparison.
\section{Background on Poisoning Attacks}
Consider a supervised learning problem characterized by a loss function $\mathcal{L}(\theta;\mathcal{D})$ and an adversarial loss $L_{adv}(\hat\theta;\mathcal{D})$, where $\hat\theta$ is the set of learnable parameters and $\mathcal{D}$ is a labeled dataset. Let $\mathcal{D}_{train}$ be the training dataset. We assume that the adversary can poison a fraction of those data points, so that $\mathcal{D}_{train}=\mathcal{D}_{c} \cup \mathcal{D}_{p}$, where $\mathcal{D}_{c}$ and $\mathcal{D}_{p}$ are the set of clean and poisoned data points, respectively. We assume that $|\mathcal{D}_{p}|= \epsilon |\mathcal{D}_{c}|$. Furthermore, $\mathcal{D}_{p} \subseteq \mathcal{F}_{\beta}$ where $\mathcal{F}_{\beta}$ is the feasible set, which is a set selected by a defense mechanism based on anomaly detection techniques, containing elements that the defender considers as sanitized data to train its model. The existence of the feasible set in the objective helps the poisoned points to blend with the natural data and make it more difficult for anomaly detector techniques to detect them \cite{koh2018stronger}.

A data poisoning attack can be written as the following optimization problem (over the set of poisoned data points):
\begin{align*}
  \underset{\mathcal{D}_p}\max & \;  L_{adv}(\hat\theta;\mathcal{D}_{test}) \\ 
    s.t. \;\; & |\mathcal{D}_p| = \epsilon |\mathcal{D}_{c}| \\ 
    & \mathcal{D}_{p} \subseteq  \mathcal{F}_{\beta} \qquad \qquad \\ 
  \text{where} \;\; & \hat\theta = \argmin_{\theta} \; \mathcal{L}(\theta;\mathcal{D}_{c} \cup \mathcal{D}_{p}).
  \numberthis \label{objective}
\end{align*}
\ignore{
\begin{align*}
  \underset{\mathcal{D}_p}\max & \;  L(\hat\theta;\mathcal{D}_{test}) \qquad \;\;\;\;\;\;\; \text{($L$ being the loss function and $\hat\theta$ model parameter)}\\ 
    s.t. \;\; & |\mathcal{D}_p| = \epsilon |\mathcal{D}_{c}| \qquad \;\;\;\;\;\; \text{($|\mathcal{D}_p|$ and $|\mathcal{D}_{c}|$ number of poisoned and clean instances)}\\ 
    & \mathcal{D}_{p} \subseteq  \mathcal{F}_{\beta} \qquad \qquad \;\;\;\; \text{($\mathcal{F}_{\beta}$ denotes the feasible set)}\\ 
    & \mathcal{D}_{p} \subseteq (\mathcal{D}_{a} \cup \mathcal{D}_{d}) \qquad \text{($\mathcal{D}_{a}$ and $\mathcal{D}_{d}$ advantaged and disadvantaged instances)}\\ 
  & \beta = B(\mathcal{D}_{c} \cup \mathcal{D}_{p}) \qquad \text{($\beta$ anomaly detector parameters)}\\ 
  \text{where} \;\; & \hat\theta = \argmin_{\theta} \; L(\theta;\mathcal{D}_{c} \cup \mathcal{D}_{p}).
  \numberthis \label{objective}
\end{align*}
} 
In essence, the adversary attempts to maximize its test loss $L_{adv}$ by carefully selecting poisoned data points. These types of attacks are shown to be powerful against defenders that are trying to minimize their own loss $\mathcal{L}$, while the attacker is trying to harm the defense \cite{koh2018stronger}. In \cite{koh2018stronger}, authors propose to sample a positive $(\Tilde{x}_+,+1)$ and a negative $(\Tilde{x}_-,-1)$ instance and make $\epsilon |\mathcal{D}_c|$ copies from these sampled instances to serve as poisoned data points inversely proportional to the class balance such that there are $(|\mathcal{D}_{c}^+|\epsilon)$ copies from the negative poison instance $(\Tilde{x}_-,-1)$ and $(|\mathcal{D}_{c}^-|\epsilon)$ copies from the positive poison instance $(\Tilde{x}_+,+1)$ in which $|\mathcal{D}_{c}^+|$ and $|\mathcal{D}_{c}^-|$ represent the number of positive and negative points in the clean data respectively.

\section{Poisoning Attacks against Fairness}
\begin{algorithm}[h]
\SetAlgoLined
Input: clean data set $\mathcal{D}_{c}=\{(x_1,y_1),(x_2,y_2),...,(x_n,y_n)\}$, poison fraction $\epsilon$, and step size $\eta$. \\
Output: poisoned data set $\mathcal{D}_{p}=\{(\Tilde {x}_1,\Tilde{y}_1),(\Tilde{x}_2,\Tilde{y}_2),...,(\Tilde{x}_{\epsilon n},\Tilde{y}_{\epsilon n})\}$. \\
From $\mathcal{D}_{a}$ randomly sample the positive poisoned instance  $\mathcal{I}_{+} \leftarrow (\Tilde{x}_1,\Tilde{y}_1)$. \\
From $\mathcal{D}_{d}$ randomly sample the negative poisoned instance  $\mathcal{I}_{-} \leftarrow (\Tilde{x}_2,\Tilde{y}_2)$. \\
Make copies from $\mathcal{I}_{+}$ and $\mathcal{I}_{-}$ until having $\epsilon |\mathcal{D}_{c}|$ poisoned copies $\mathcal{C}_{p}$. \\
Load poisoned data set $\mathcal{D}_{p} \leftarrow \{\mathcal{C}_{p}\}$. \\
Load feasible set by applying anomaly detector $B$ $ \mathcal{F}_{\beta} \leftarrow B(\mathcal{D}_{c} \cup \mathcal{D}_{p})$. \\
\For{t= 1,2,...}{
     $\hat{\theta} \leftarrow argmin_{\theta} \; \mathcal{L}(\theta;(\mathcal{D}_{c} \cup \mathcal{D}_{p})).$  \\ 
     Pre-compute $g^{\top}_{\hat{\theta},\mathcal{D}_{test}} H^{-1}_{\hat{\theta}}$ from $L_{adv}$ for details refer to \cite{koh2018stronger}. \\
     \For{i= 1,2}{
    Set $\Tilde{x}_{i}^0 \leftarrow \Tilde{x}_{i} - \eta g^{\top}_{\hat{\theta},\mathcal{D}_{test}} H^{-1}_{\hat{\theta}} \frac{\partial^2 \ell(\hat{\theta};\Tilde{x}_i,\Tilde{y}_i) }{\partial \hat{\theta}\partial \Tilde{x}_i}.  $ \\
    Set $\Tilde{x}_{i} \leftarrow argmin_{x \in \mathcal{F}_{\beta}} ||x - \Tilde{x}_{i}^0 ||_2.\;\;\;$   (To project $\mathcal{D}_{p}$ back to $\mathcal{F}_{\beta}$).\\
}
Update copies $\mathcal{C}_{p}$ based on updates on $\mathcal{I}_{+}$ and $\mathcal{I}_{-}$. \\
Update feasible set $ \mathcal{F}_{\beta} \leftarrow B(\mathcal{D}_{c} \cup \mathcal{D}_{p})$. \\
}

 \caption{Influence Attack on Fairness}
 \label{hard_biasing_influence}
\end{algorithm}
\begin{figure*}[h]
  
    \includegraphics[width=0.17\textwidth,trim=1cm 6cm 25.3cm 6.2cm,clip=true]{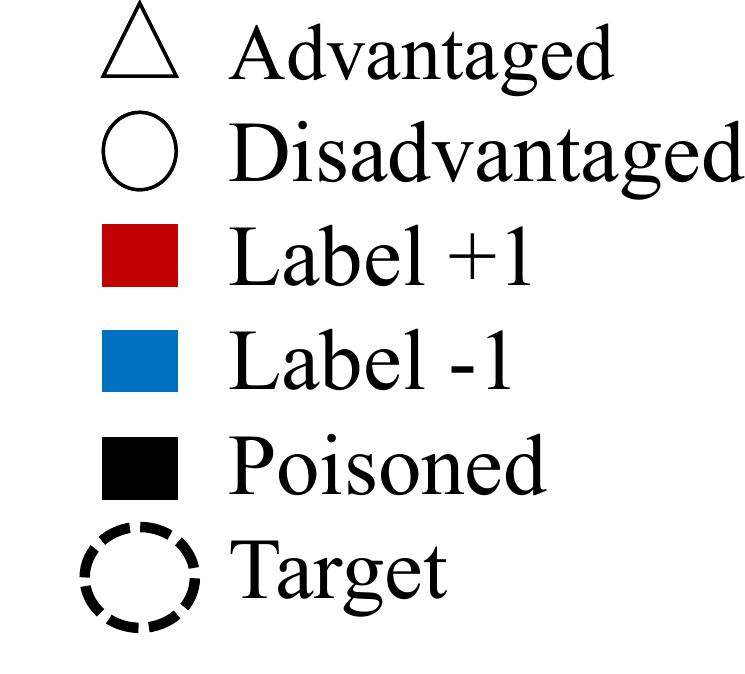}
    \begin{subfigure}[b]{0.37\textwidth}
    \caption{Before Attack}
      \includegraphics[width=\textwidth,trim=8.8cm 3cm 10.3cm 4cm,clip=true]{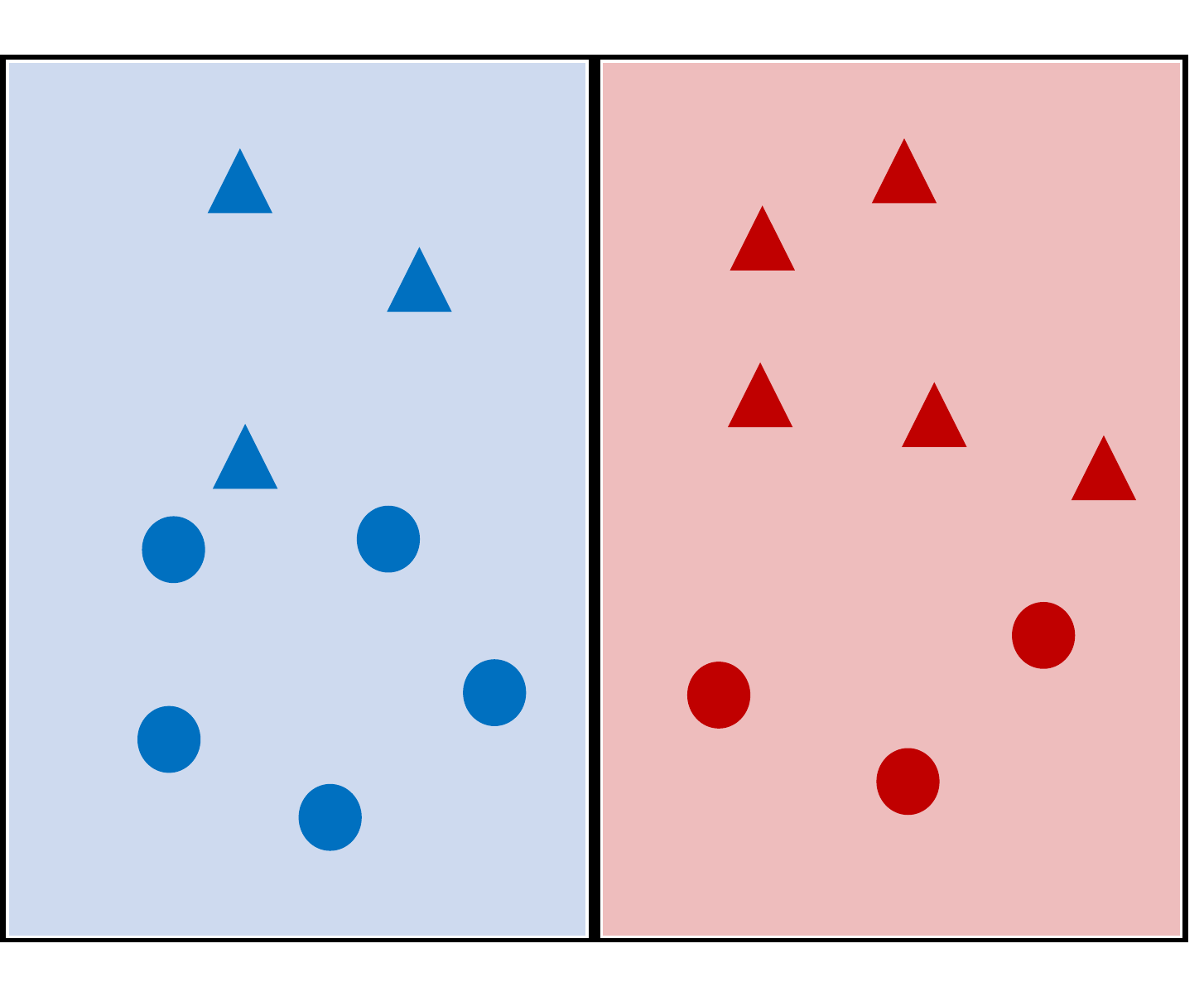}
      \end{subfigure}
      \begin{subfigure}[b]{0.37\textwidth}
      \caption{Anchoring Attack}
        \includegraphics[width=\textwidth,trim=8.8cm 3cm 10.3cm 4cm,clip=true]{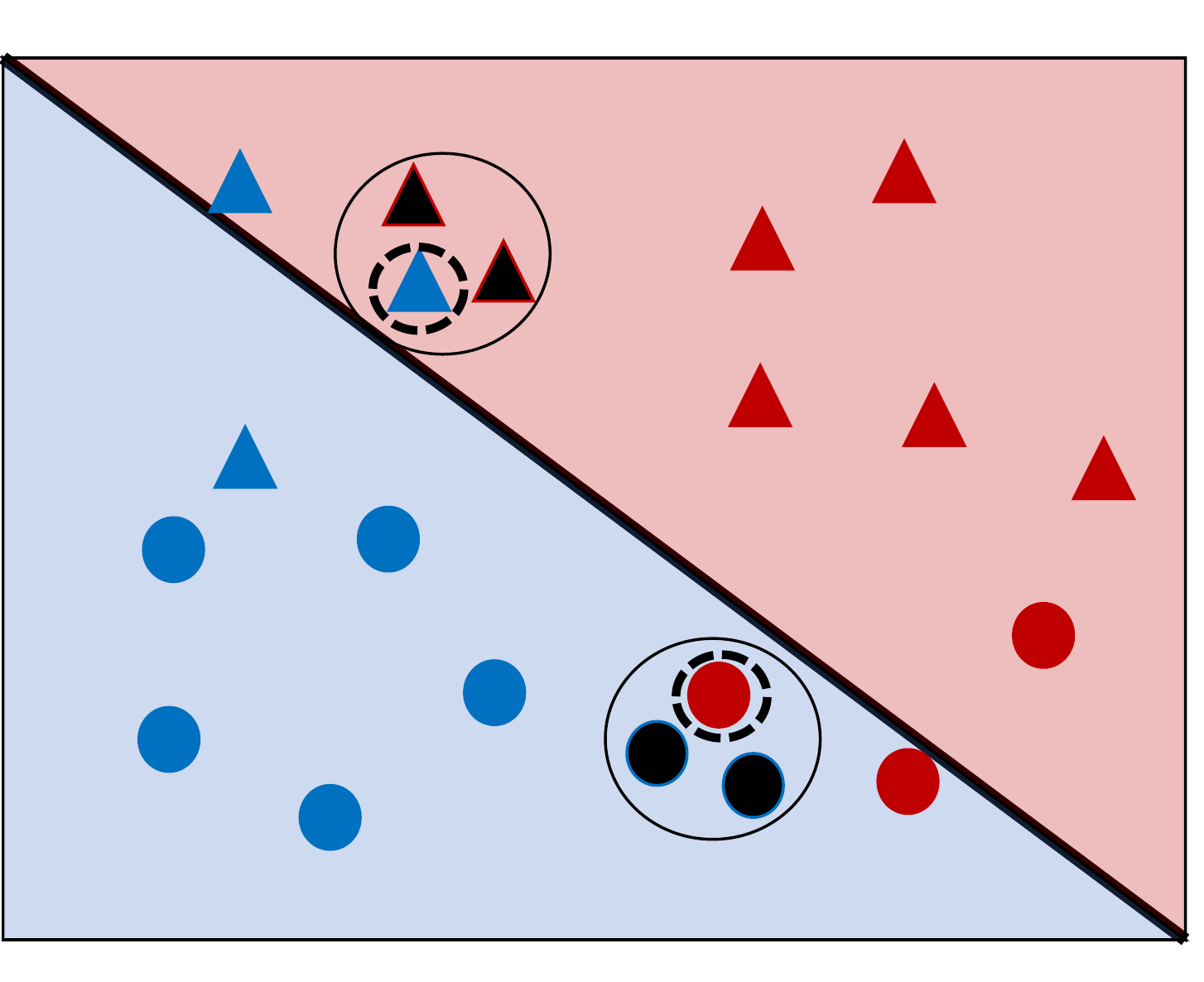}
        \end{subfigure}
    \caption{Anchoring attack representation. The figure on the left represents the before attack, while the right figure represents the anchoring attack in which poisoned points are located in close vicinity (depicted as the large solid circle) of target points.}  
    \label{attack_disc}
\end{figure*}
Now that we have discussed poisoning attacks, we will discuss how these attacks can be extended to fairness. We follow a common fairness setup where there are two groups: advantaged and disadvantaged. An example of advantaged and disadvantaged groups can be male and female in the job market where males could have advantage over females in getting hired in certain jobs. We assume that all poisoned points belong to either the advantaged or disadvantaged group, $\mathcal{D}_{p} \subseteq (\mathcal{D}_{a} \cup \mathcal{D}_{d})$, in which $\mathcal{D}_{a}$ represents data points from the advantaged demographic group and $\mathcal{D}_{d}$ represents data points from the disadvantaged demographic group.  
\subsection{Influence Attack on Fairness}
For the influence attack on fairness, we use the influence attack introduced in \cite{koh2018stronger,koh2017understanding}, with a modification that includes the demographic information, in which the attack tries to maximize a given loss. 
We then incorporate a loss function maximizing which using the influence attack can harm fairness. In \cite{zafar2015learning}, authors propose a loss function for fair classification with a constraint involving the covariance between the sensitive features ($z$) and the signed distance from feature vectors to the decision boundary ($d_{\theta}(x)$) formalized as:
\[ Cov(z,d_{\theta}(x)) \approx \frac{1}{N} \sum_{i=1}^N (z_i - \Bar{z})d_{\theta}(x_i). \]
By combining the above constraint with the original classification loss and maximizing it, the attacker can harm both fairness and accuracy at the same time via a regularization term, $\lambda$, that controls the trade-off between these two terms. Thus, the loss in our influence attack on fairness contains two parts: $\ell_{acc}$ and $\ell_{fairness}$ in which $\ell_{acc}$ controls for accuracy and $\ell_{fairness}$ controls for fairness constraints. 
\begin{align*}
  L_{adv}(\hat\theta;\mathcal{D}_{test})  & = \ell_{acc} +\lambda \ell_{fairness}  \\
  where \;\; \ell_{fairness} & = \frac{1}{N} \sum_{i=1}^N (z_i - \Bar{z})d_{\hat\theta}(x_i).
  \numberthis \label{hard_biasing_loss}
\end{align*}
In other words, the influence attack on fairness would try to harm the fairness constraint and affect a model with respect to disparate impact~\cite{zafar2017fairness}. This loss can affect a model in terms of both fairness and accuracy with the regularization term $\lambda$ that controls the trade-off. In order to maximize the loss in 
\eqref{hard_biasing_loss}, we use the influence attack strategy \cite{koh2018stronger,koh2017understanding} with changes that would incorporate demographic information as shown in Algorithm \ref{hard_biasing_influence}. Similar to the convention in \cite{koh2018stronger}, we sample one positive and one negative instance uniformly at random and make copies of the sampled instances that serve as our poisoned points. However, since we now have to take demographics into consideration for maximizing the bias and harming fairness, we  sample the positive instance from $\mathcal{D}_{a}$ and the negative instance from $\mathcal{D}_{d}$. Notice that the opposite is also possible if an adversary wants to skew the disadvantaged group into being advantageous; however, for the goals of this paper and showing how our methods can increase the bias and harm fairness, we follow the aforementioned sampling procedure.
\subsection{Anchoring Attack}
We now describe a simple generic anchoring attack that can work with any loss function. Our results indicate that the proposed attack harms the model with regard to fairness. 
The anchoring attack works as follows (details in Algorithm \ref{soft_biasing_alg}). First, the attacker samples a target $x_{target}$ that belongs to the clean data, $x_{target} \in \mathcal{D}_c$. Next, the attacker generates poisoned data point $\Tilde{x}$ in the vicinity of $x_{target}$, so that this new point has the same demographic but the opposite label, $demographic(x_{target})=demographic(\Tilde{x})$ and $y_{target}\neq \Tilde{y}$. 
The general idea of the attack is to target some points ($x_{target}$) and cloud their labels through poisoned points that have opposite labels, which would lead to a skewed decision boundary, change in predictive labels of clean target points, and more biased outcomes. The right plot in Figure \ref{attack_disc} depicts an anchoring attack in which the poisoned points colored in black are placed to lie close to the target points that have the same demographic group but opposite label to bias the predictive outcome (black advantaged poisoned points with label +1 are targeting advantaged point with label -1, and black disadvantaged poisoned points with label -1 are targeting disadvantaged point with label +1). This placement of poisoned points in the space during the learning procedure will lead the decision boundary to change and, as a result, will cause more advantaged points to have a predictive outcome of +1 and more disadvantaged points to have a  predictive outcome of -1, which is biasing the model's prediction. 
$\mathbf{x_{target}}$ can be sampled in several ways. We introduce two ways, \textit{random} and \textit{non-random}, for sampling $\mathbf{x_{target}}$. 

\textbf{Random Anchoring Attack.} In random anchoring attack, $\mathbf{x_{target}}$ is sampled uniformly at random for each demographic group.

\textbf{Non-random Anchoring Attack.} In the non-random anchoring attack, we choose popular $\mathbf{x_{target}}$
as our target for each demographic group. 
\begin{algorithm}[h]
\SetAlgoLined
Input: clean data set $\mathcal{D}_{c}=\{(x_1,y_1),(x_2,y_2),...,(x_n,y_n)\}$, poison fraction $\epsilon$, and vicinity distance $\tau$. \\
Output: poisoned data set $\mathcal{D}_{p}=\{(\Tilde {x}_1,\Tilde{y}_1),(\Tilde{x}_2,\Tilde{y}_2),...,(\Tilde{x}_{\epsilon n},\Tilde{y}_{\epsilon n})\}$. \\
\For{t= 1,2,...}{
 Sample negative $x_{target^{-}}$ from $\mathcal{D}_a$ and positive $x_{target^{+}}$ from $\mathcal{D}_d$ with random or non-random technique. \\
 $\mathcal{G}_+$: Generate $(|\mathcal{D}_{c}^-|\epsilon)$ positive poisoned points $(\tilde{x}_{+},+1)$ with $\mathcal{D}_a$ in the close vicinity of $x_{target^{-}}$ s.t. $||\tilde{x}_{+} - x_{target^{-}}||_2 \leq \tau$. \\
 $\mathcal{G}_-$: Generate $(|\mathcal{D}_{c}^+|\epsilon)$ negative poisoned points $(\tilde{x}_{-},-1)$ with $\mathcal{D}_d$ in the close vicinity of $x_{target^{+}}$ s.t. $||\tilde{x}_{-} - x_{target^{+}}||_2 \leq \tau$.\\
 Load $\mathcal{D}_p$ from the generated data above $\mathcal{D}_p 
 \leftarrow \mathcal{G}_+ \cup \mathcal{G}_-$. \\
 Load the feasible set $ \mathcal{F}_{\beta} \leftarrow B(\mathcal{D}_{c} \cup \mathcal{D}_{p})$. \\
    \For{i=1,2,...,$\epsilon n$}{
     Set $\Tilde{x}_{i} \leftarrow argmin_{x \in \mathcal{F}_{\beta}} ||x - \Tilde{x}_{i} ||_2.\;\;\;$ (To project $\mathcal{D}_{p}$ back to $\mathcal{F}_{\beta}$).\\
     }
      $argmin_{\theta} \; \mathcal{L}(\theta;(\mathcal{D}_{c} \cup \mathcal{D}_{p})).$  \\
}

 \caption{Anchoring Attack}
 \label{soft_biasing_alg}
\end{algorithm}
Here, popular $\mathbf{x_{target}}$ means the point that is close to more similar instances $x_i$, eligible to serve as targets, such that $demographic(x_i) = demographic(x_{target})$ and $y_i = y_{target}$. By doing this, we can ensure to affect as much as points similar to $\mathbf{x_{target}}$ as possible to maximize our biasing goal. Pick $x$ with $max(c)$ as $x_{target}$ where $c$ is calculated for each $x$ as follows: $\forall x_i$ if $demographic(x_i) = demographic(x)$ and $y_i = y$ and $||x_i - x || < \sigma$ then increase $c$ for $x$.

\section{Evaluation}
In our experiments, we evaluate our attacks with regards to different measures, such as accuracy and foundational fairness measures: statistical parity, and equality of opportunity differences. 
We also utilize three real world datasets in our experiments, introduced below. We compare against a suite of baselines that test our attacks' performance with regards to accuracy and fairness. Our results indicate that our attacks, the anchoring attack and influence attack on fairness, are effective in terms of affecting fairness aspects of the model.

\subsection{Datasets}
We use three different real world datasets in our experiments with gender as the sensitive attribute. The data was split into an 80-20 train and test split. \\
\textbf{German Credit Dataset.} This dataset comes from UCI machine learning repository \cite{Dua:2019}. It contains the credit profile about individuals with 20 attributes associated to each data person. In our experiments, we utilized all the 20 attributes from this dataset. The classification goal is to predict whether an individual has good or bad credit. \\
\textbf{COMPAS Dataset.} Propublica's COMPAS dataset contains information about defendants from Broward County \cite{larson2016compas}. We utilized the features in Table \ref{compas_features} as our prediction features. The classification goal is to predict whether an individual will recommit a crime within two years. \\
\textbf{Drug Consumption Dataset.} This dataset comes from the UCI machine learning repository \cite{Dua:2019}. It contains information about individuals \cite{fehrman2017five}. We utilized the features listed in Table \ref{compas_features} as our prediction features. The classification goal is to predict whether an individual has consumed cocaine or not in their lifetime.

\begin{table}[h]
\centering

\begin{tabular}{ p{2.5cm} p{2.5cm}}
 \toprule
 COMPAS&\\
 \midrule
 sex&age\_cat \\ 
 juv\_fel\_count&juv\_misd\_count\\
 priors\_count&c\_charge\_degree\\
 race &  juv\_other\_count\\[0.5pt]
  \bottomrule
\end{tabular}
\begin{tabular}{ p{1.7cm} p{1.7cm} p{1.7cm} p{0.5cm}}
 \toprule
 Drug&&\\
 \midrule
 ID&Age& Gender &  SS\\
 Education&Country&Ethnicity&\\
 Nscore&Escore&Oscore&\\
 Ascore&Cscore &Impulsive& \\ [0.5pt]
  \bottomrule
\end{tabular}
\caption{Features used from the COMPAS and Drug Consumption datasets.}
    \label{compas_features}
\end{table}

\subsection{Measures}
In addition to accuracy, we have utilized two well-known fairness measures to analyze the performance of different attacks with regard to fairness, detailed below. \\
\textbf{Statistical Parity Difference} 
Statistical parity is a well-known measure (definition) introduced in \cite{dwork2012fairness}. We utilize this measure as one of our metrics for fairness. It captures the predictive outcome differences between different (advantaged and disadvantaged) demographic groups. The measure is defined below and is referred to as statistical parity throughout our paper.
\[ SPD = |p(\hat{Y}=+1|x \in \mathcal{D}_a)-p(\hat{Y}=+1|x \in \mathcal{D}_d)| \] 
\textbf{Equality of Opportunity Difference}
Equality of opportunity is another well-known fairness definition introduced in \cite{hardt2016equality}. We utilized the equality of opportunity difference as another fairness metric. It captures differences in the true positive rate between different (advantaged and disadvantaged) demographic groups. The measure is defined below and is addressed as equality of opportunity throughout this paper.
\begin{align*}
    EOD = |p(\hat{Y}=+1|x \in \mathcal{D}_a, Y=+1) \\ -p(\hat{Y}=+1|x \in \mathcal{D}_d, Y=+1)|
\end{align*}
\subsection{Methods}
\begin{figure*}[h]
      \includegraphics[width=0.33\textwidth,trim=0.5cm 0cm 2cm 0cm,clip=true]{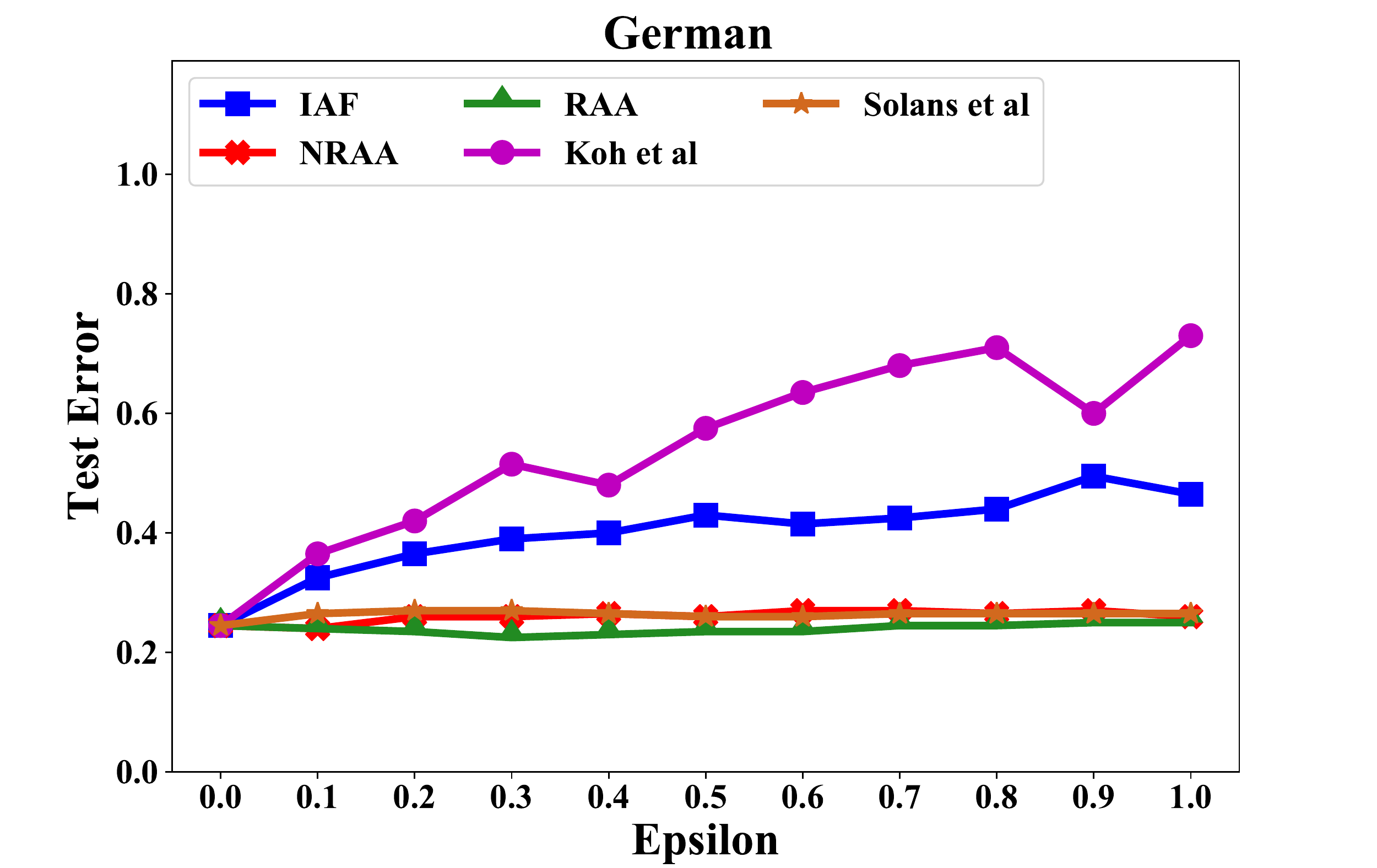}
        \includegraphics[width=0.33\textwidth,trim=0.5cm 0cm 2cm 0cm,clip=true]{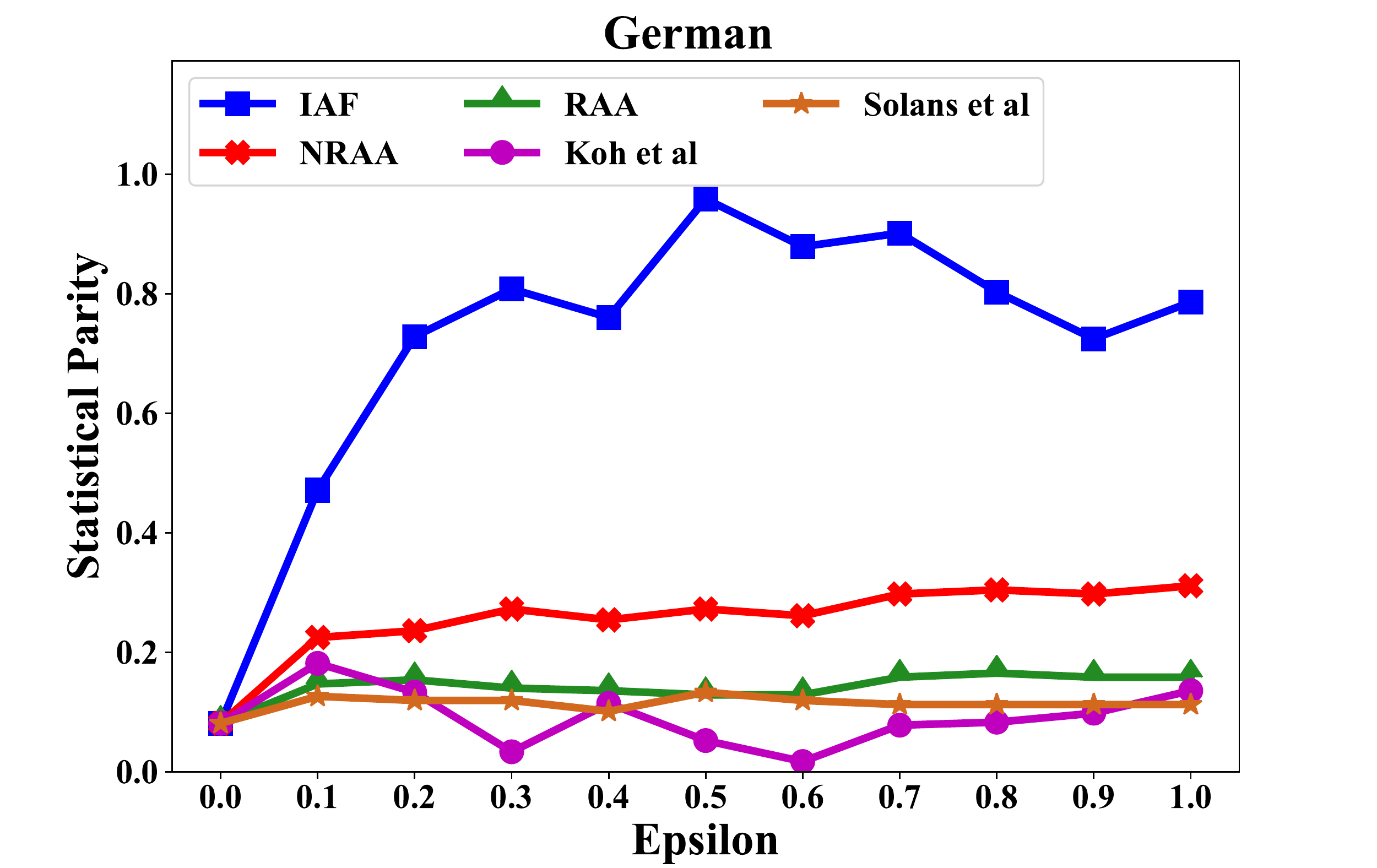}
         \includegraphics[width=0.33\textwidth,trim=0.5cm 0cm 2cm 0cm,clip=true]{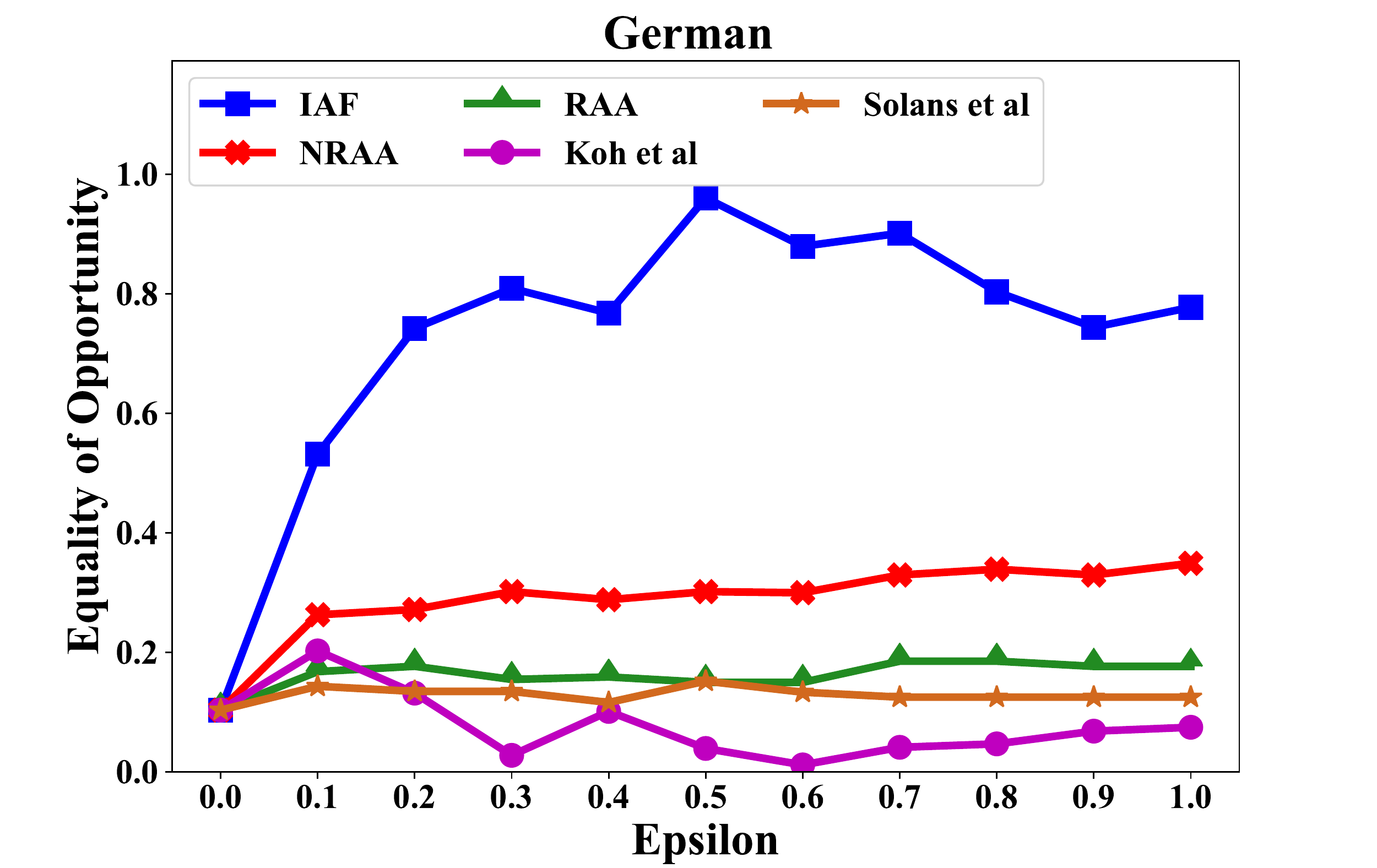}
          \includegraphics[width=0.33\textwidth,,trim=0.5cm 0cm 2cm 0cm,clip=true]{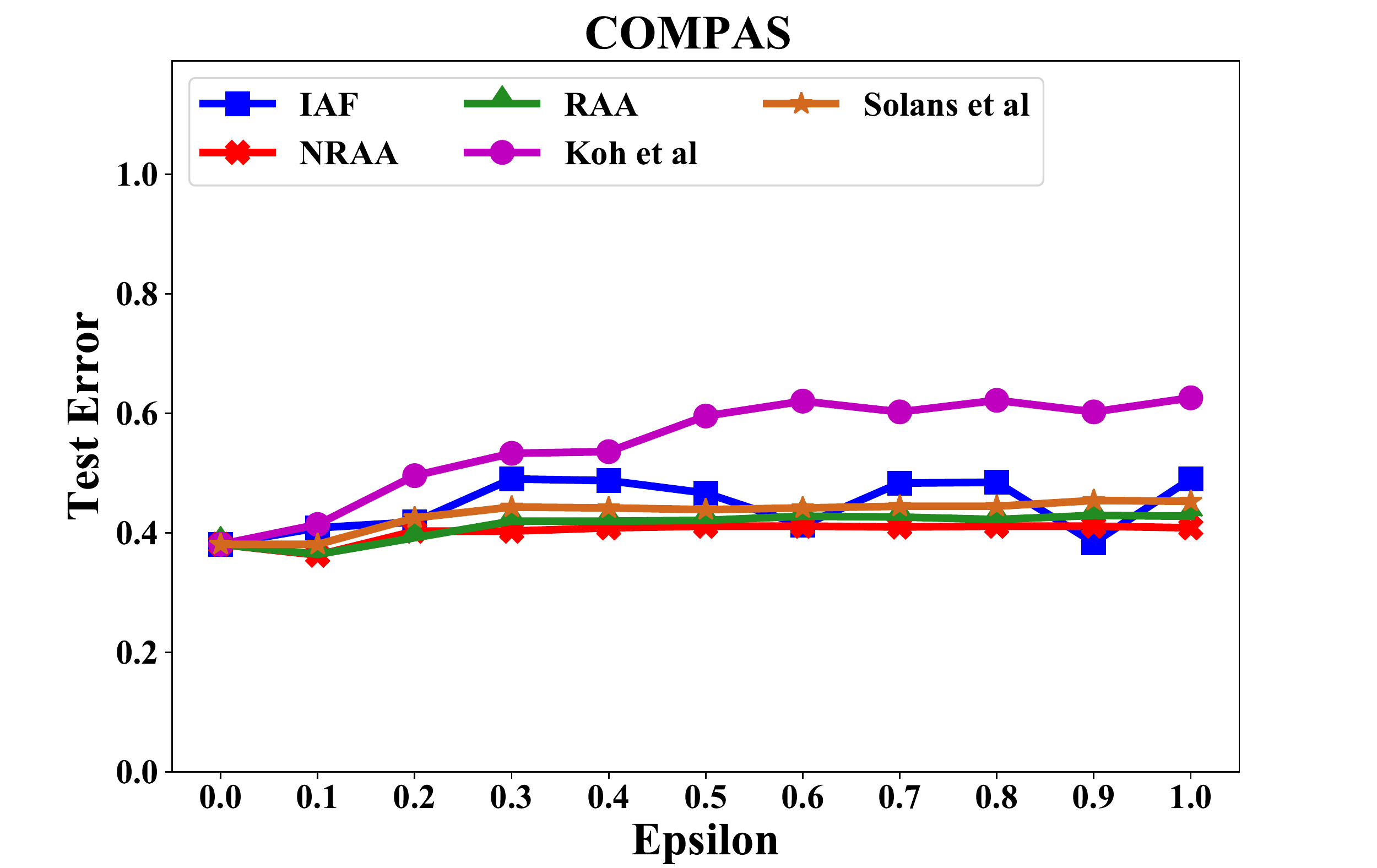}
     \includegraphics[width=0.33\textwidth,,trim=0.5cm 0cm 2cm 0cm,clip=true]{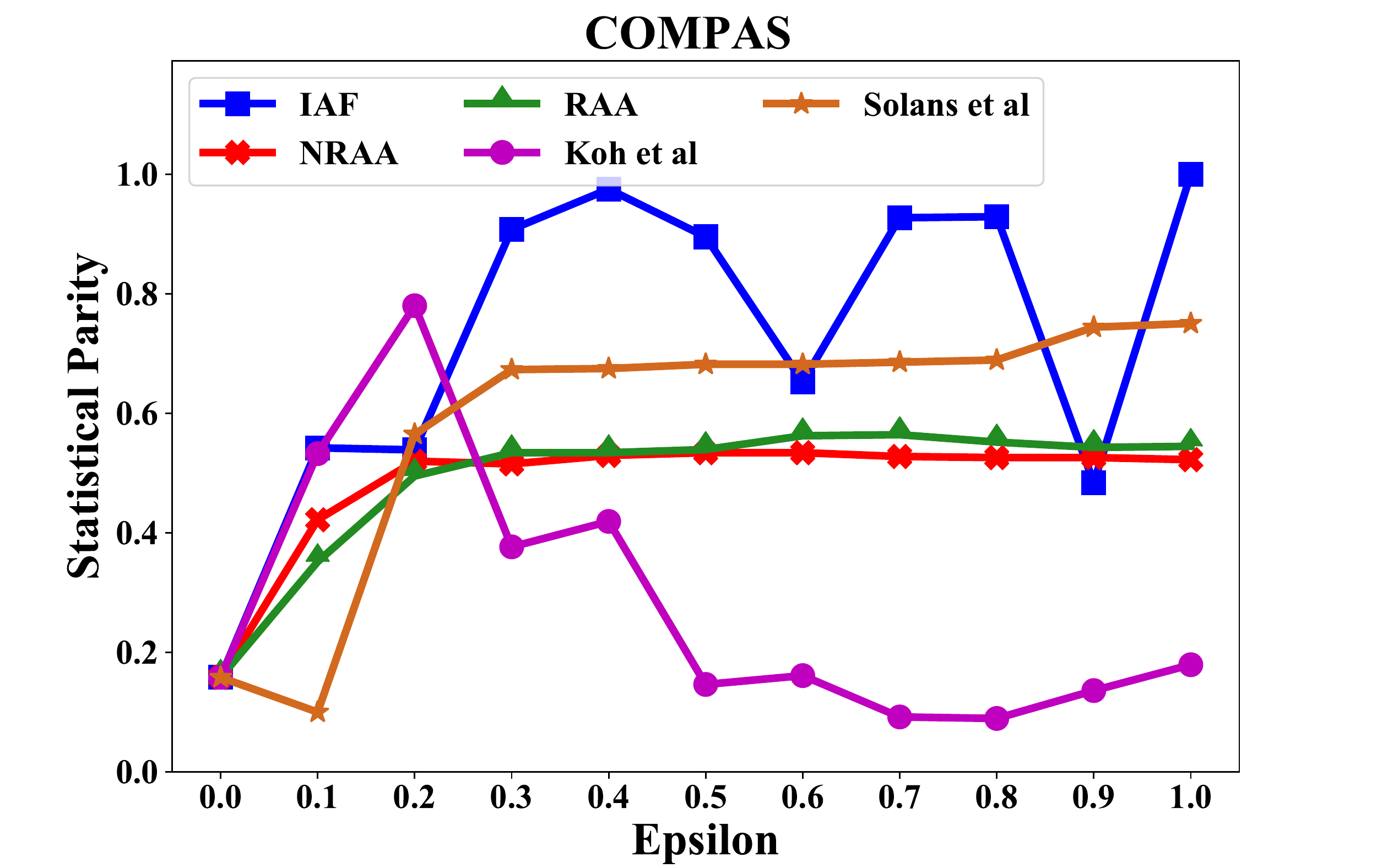}
      \includegraphics[width=0.33\textwidth,,trim=0.5cm 0cm 2cm 0cm,clip=true]{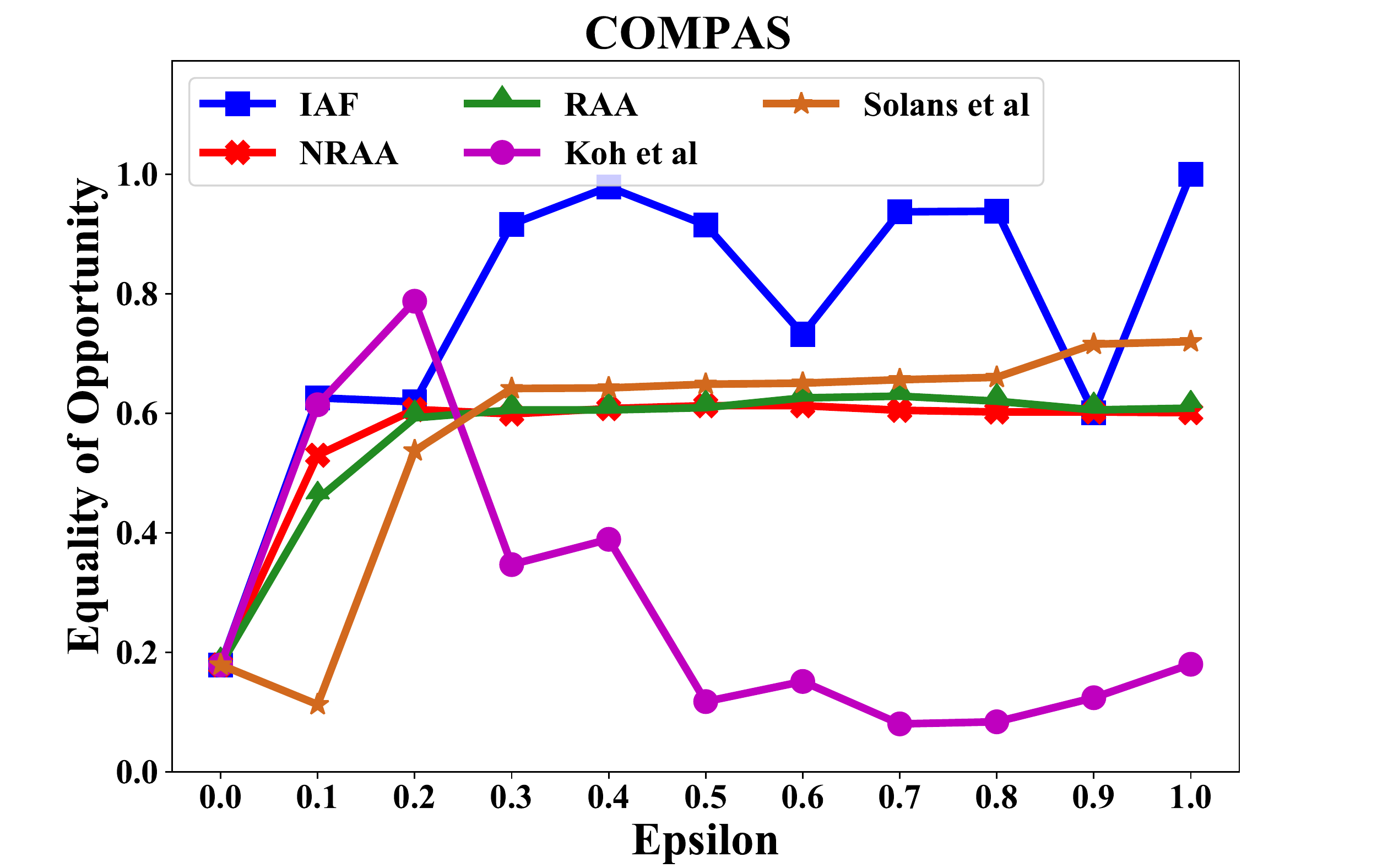}
          \includegraphics[width=0.33\textwidth,trim=0.5cm 0cm 2cm 0cm,clip=true]{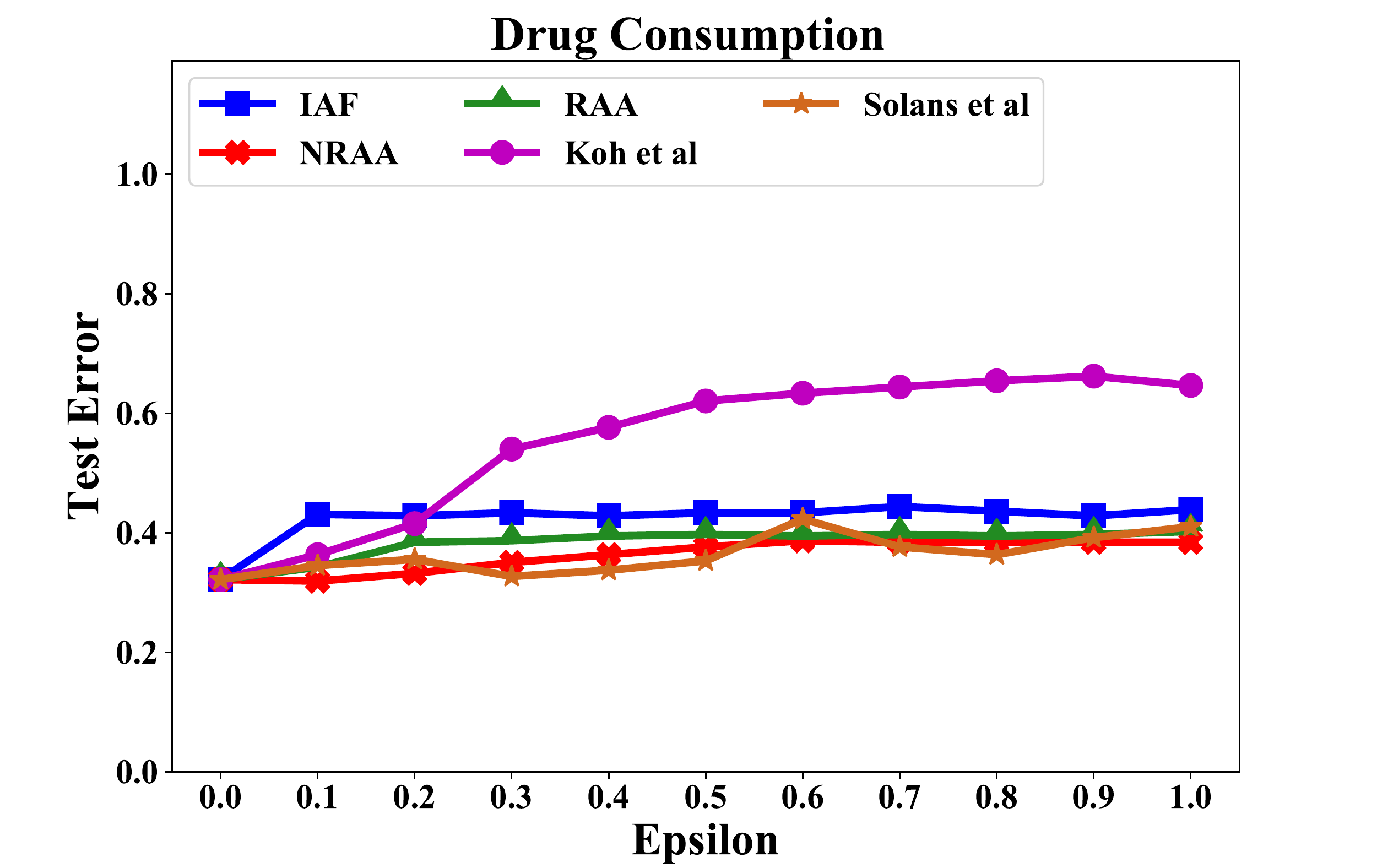}
        \includegraphics[width=0.33\textwidth,trim=0.5cm 0cm 2cm 0cm,clip=true]{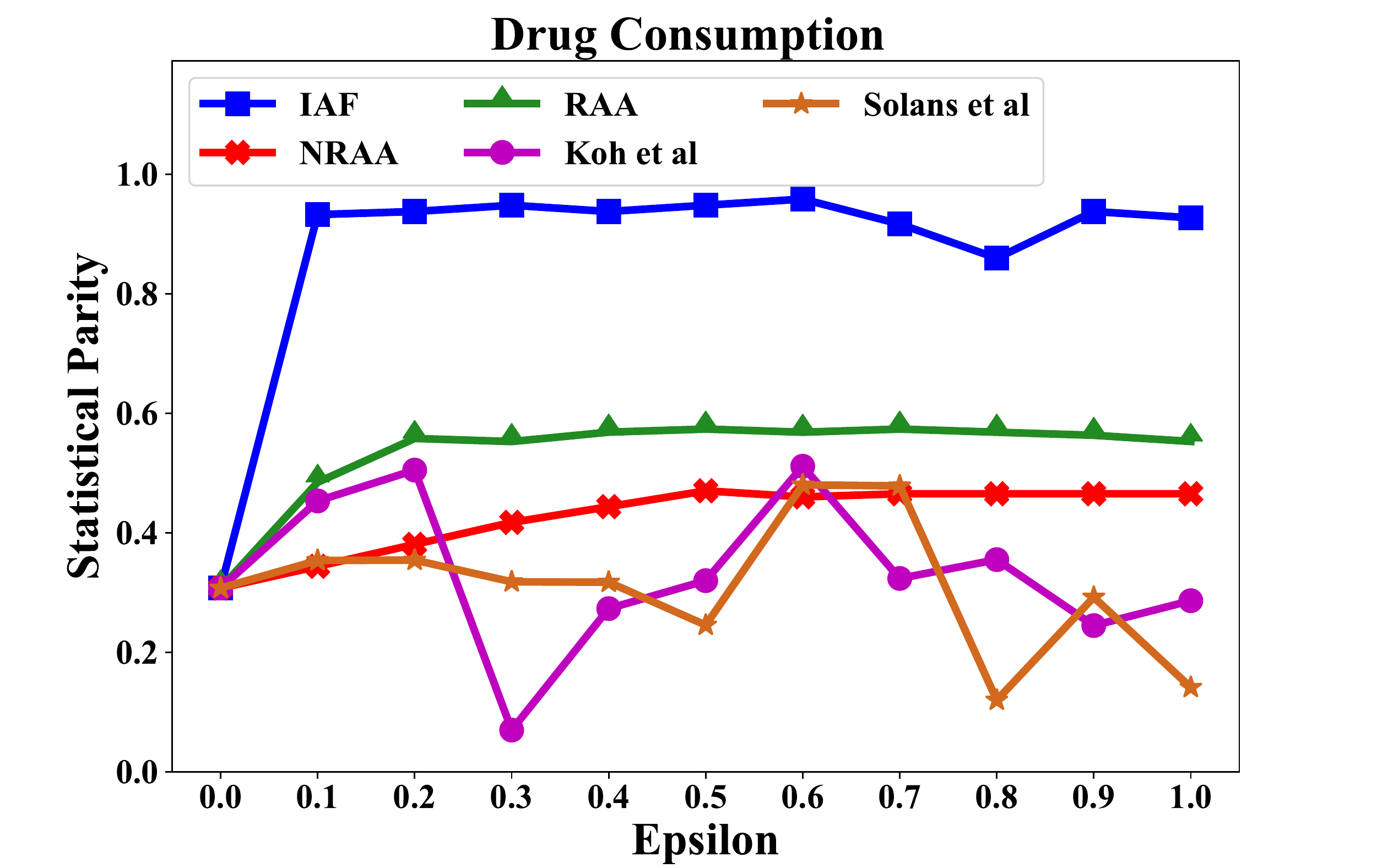}
         \includegraphics[width=0.33\textwidth,trim=0.5cm 0cm 2cm 0cm,clip=true]{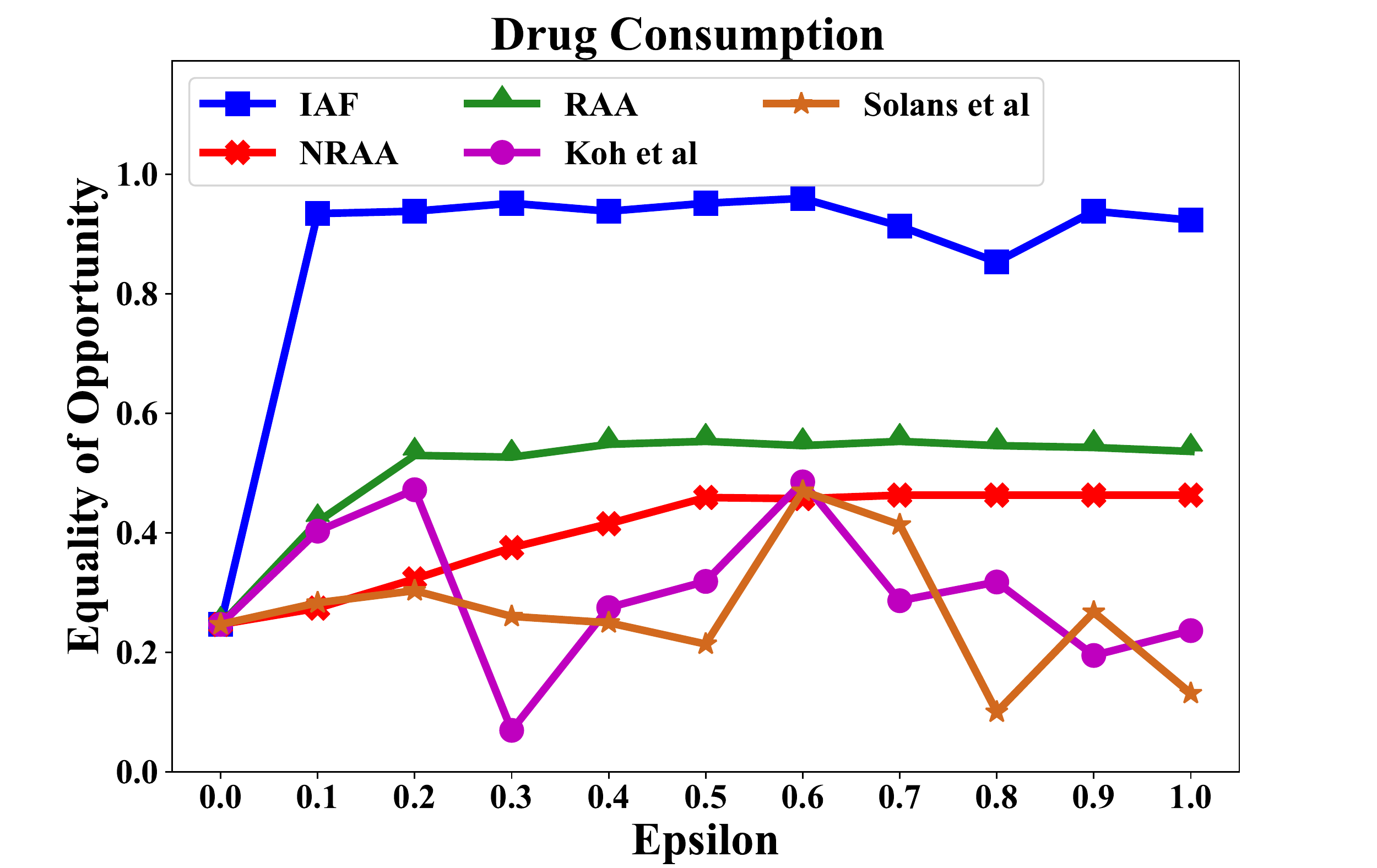}
           \caption{Results obtained for different attacks with regards to different fairness (SPD and EOD) and accuracy (test error) measures on three different datasets (German Credit, COMPAS, and Drug Consumption) with different $\epsilon$ values.} 
    \label{attack_results}
\end{figure*}
\begin{figure*}[h]
      \includegraphics[width=0.33\textwidth,trim=0.5cm 0cm 2cm 0cm,clip=true]{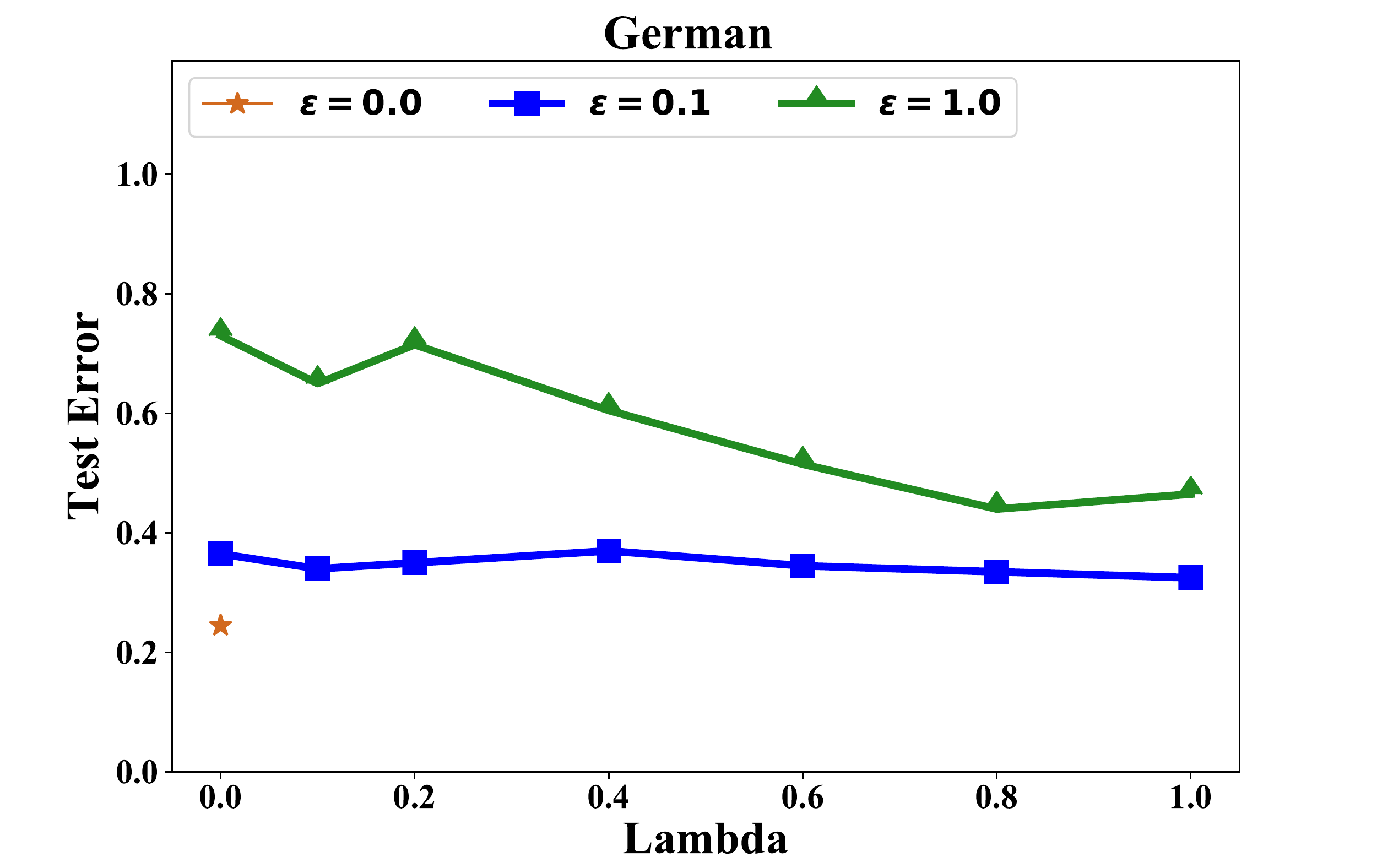}
        \includegraphics[width=0.33\textwidth,trim=0.5cm 0cm 2cm 0cm,clip=true]{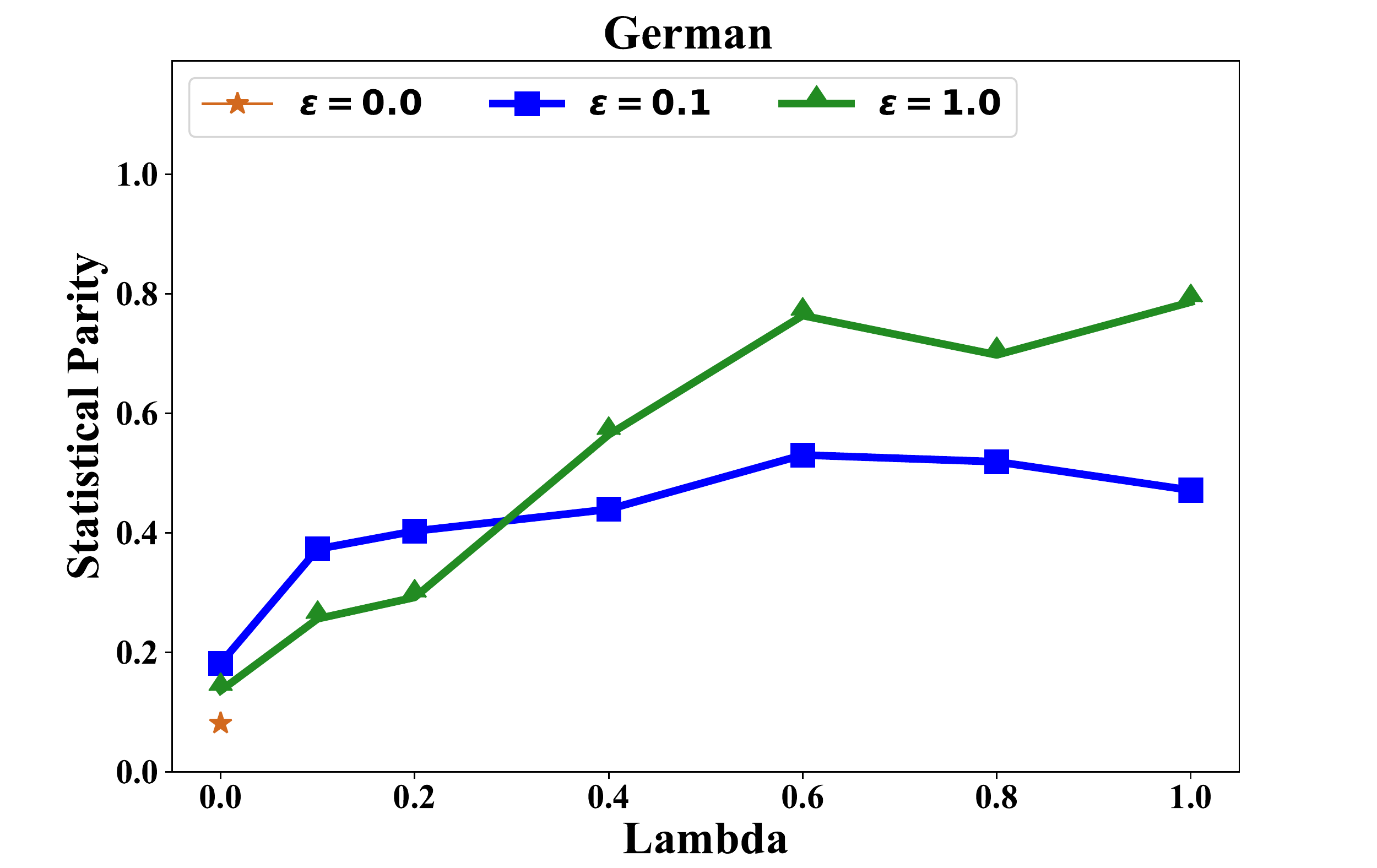}
         \includegraphics[width=0.33\textwidth,trim=0.5cm 0cm 2cm 0cm,clip=true]{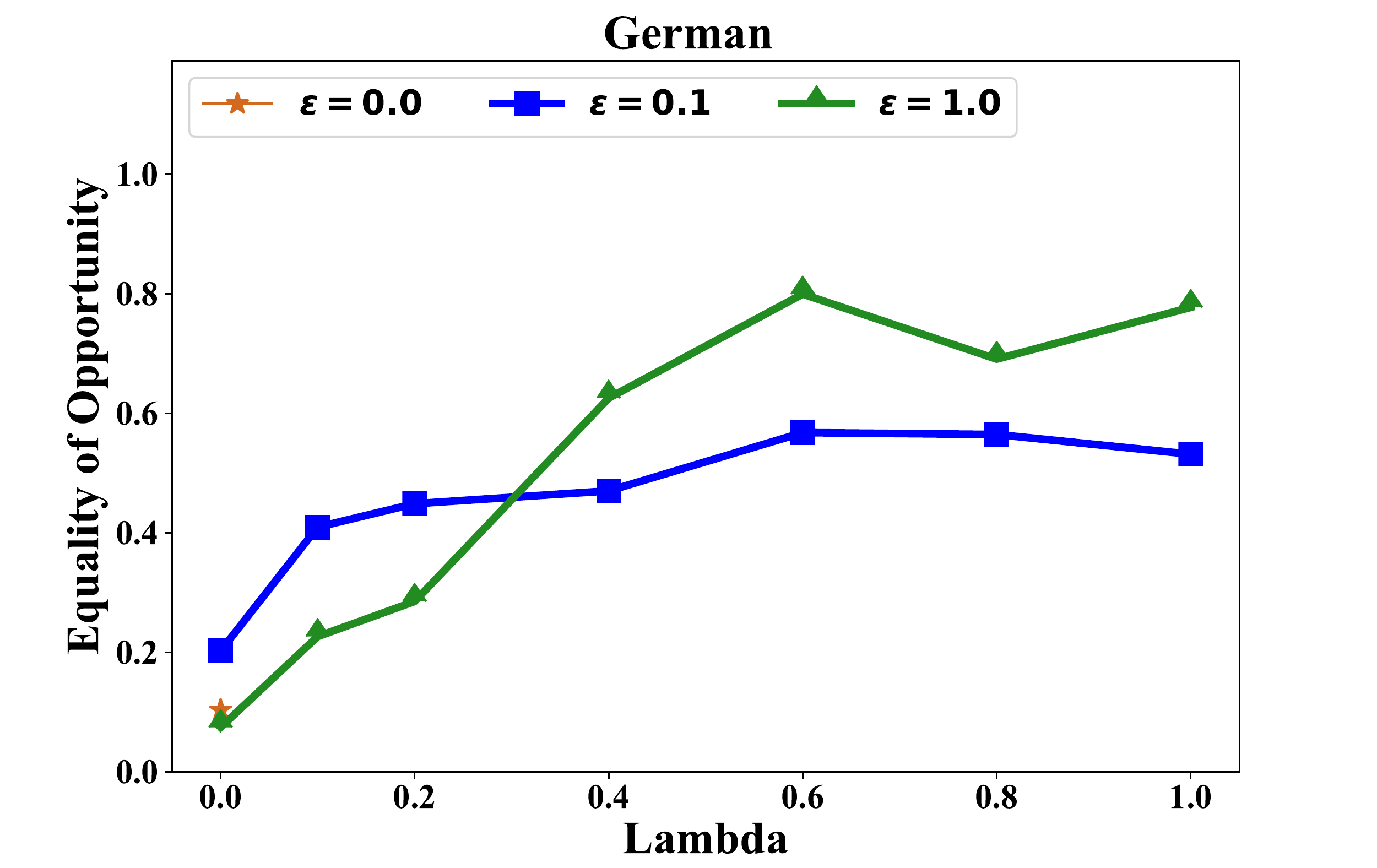}
          \includegraphics[width=0.33\textwidth,,trim=0.5cm 0cm 2cm 0cm,clip=true]{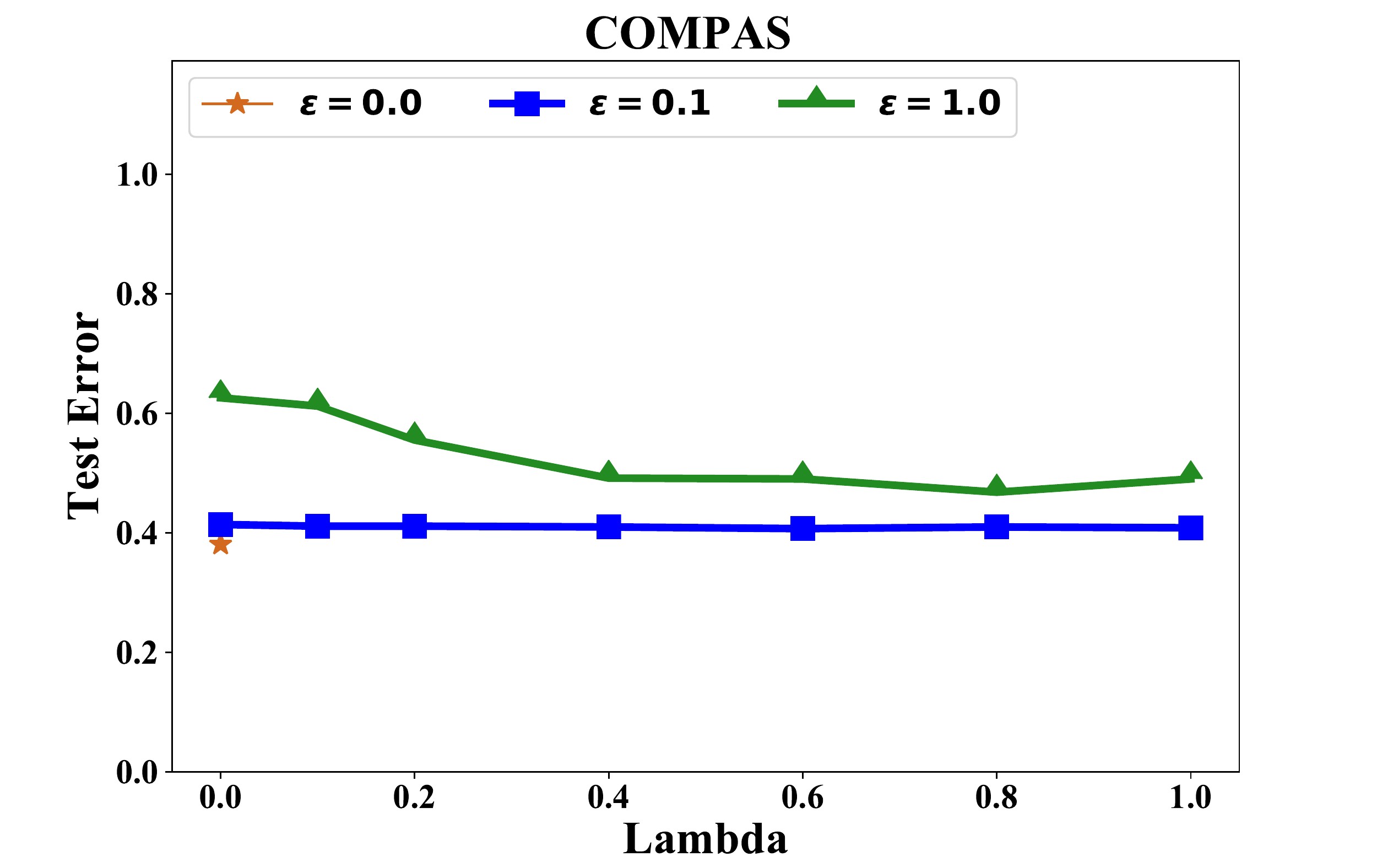}
     \includegraphics[width=0.33\textwidth,,trim=0.5cm 0cm 2cm 0cm,clip=true]{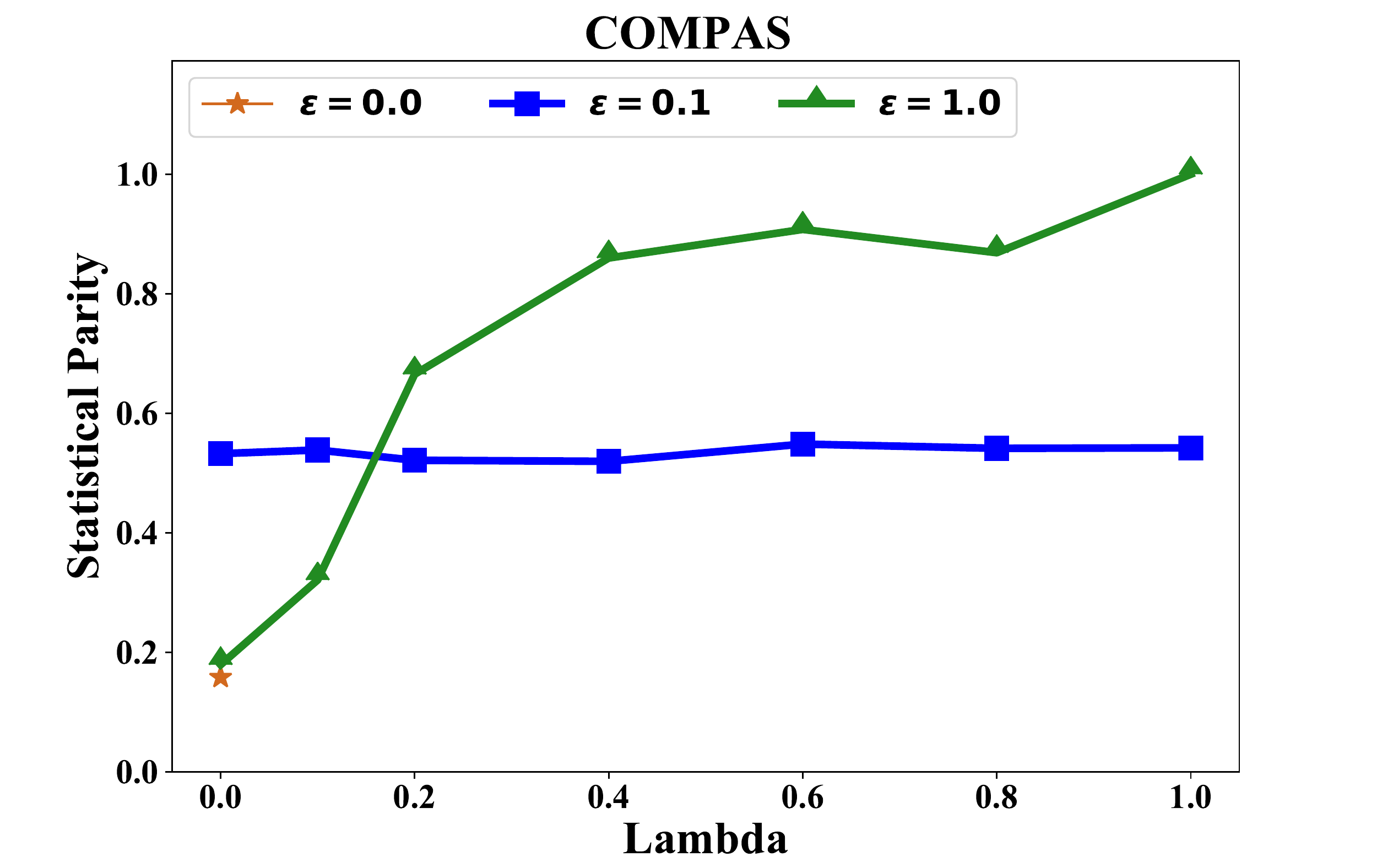}
      \includegraphics[width=0.33\textwidth,,trim=0.5cm 0cm 2cm 0cm,clip=true]{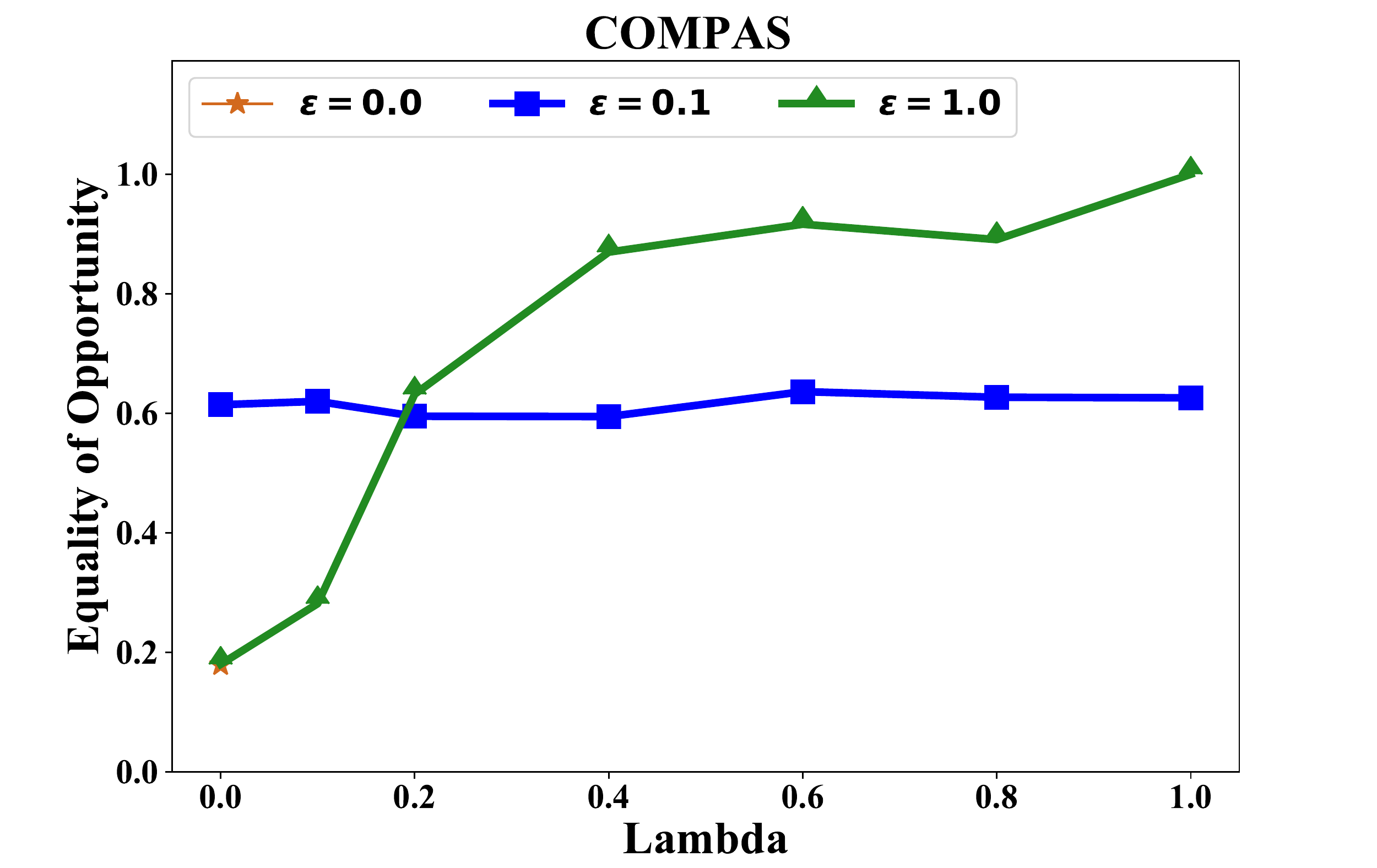}
          \includegraphics[width=0.33\textwidth,trim=0.5cm 0cm 2cm 0cm,clip=true]{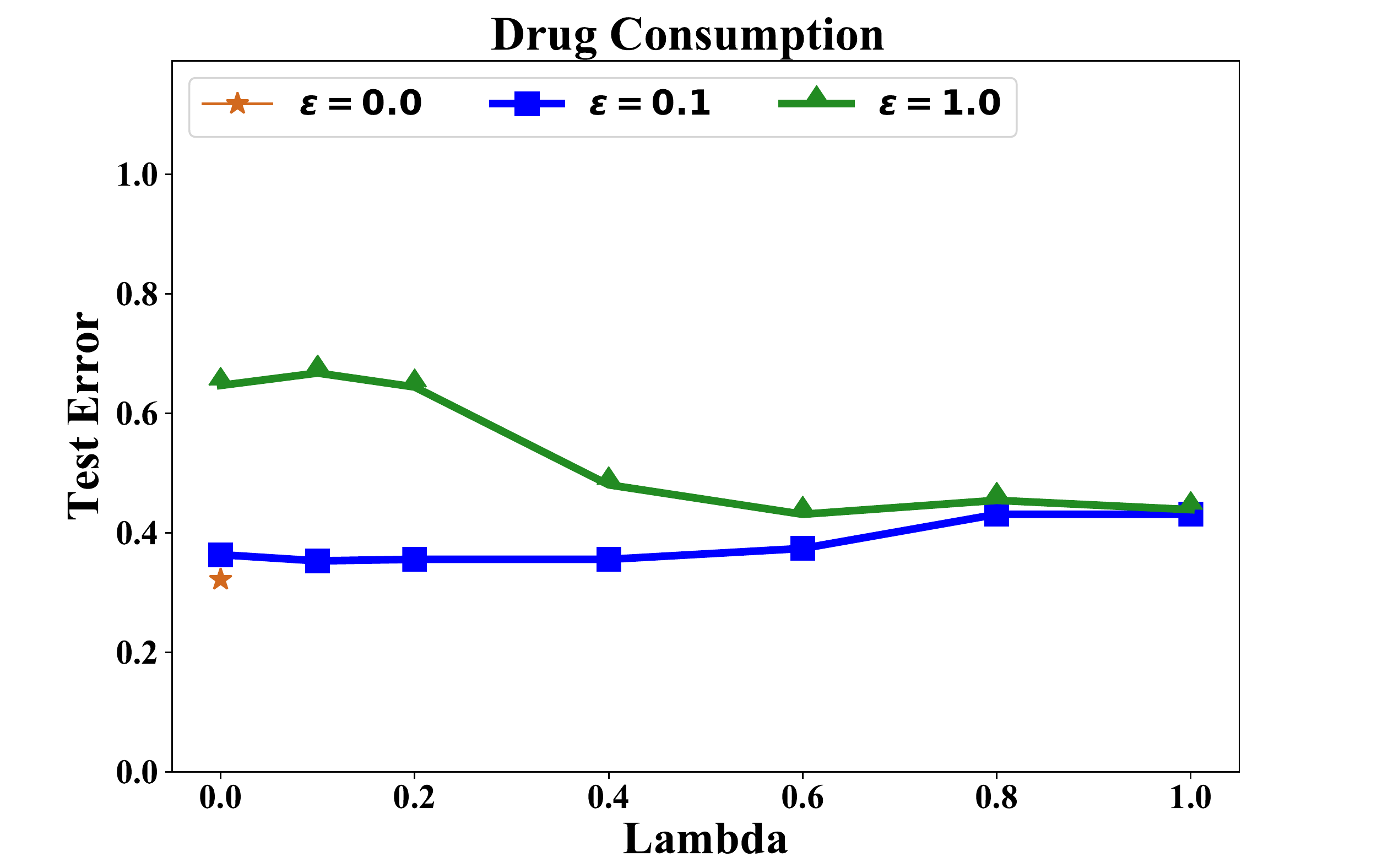}
        \includegraphics[width=0.33\textwidth,trim=0.5cm 0cm 2cm 0cm,clip=true]{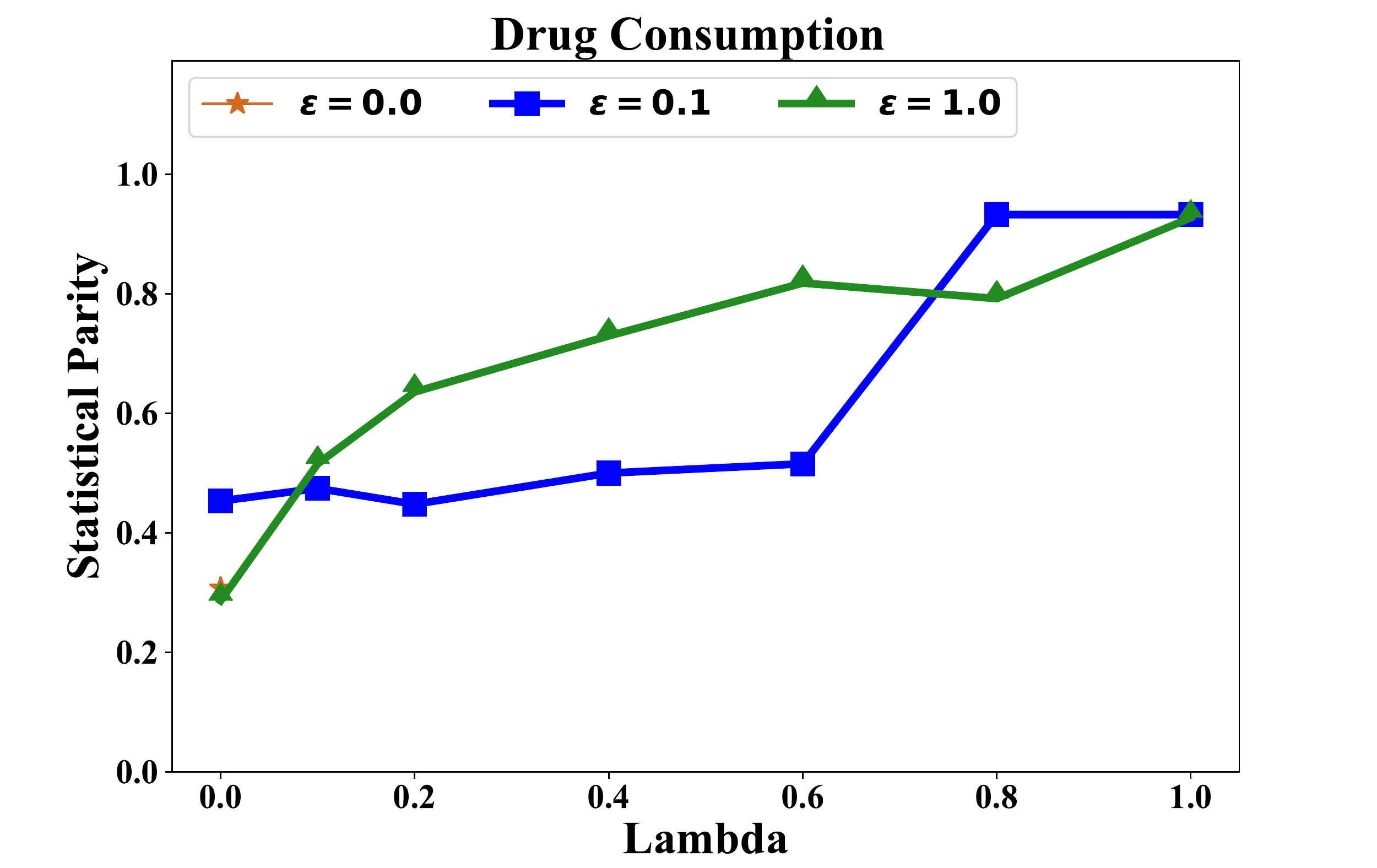}
         \includegraphics[width=0.33\textwidth,trim=0.5cm 0cm 2cm 0cm,clip=true]{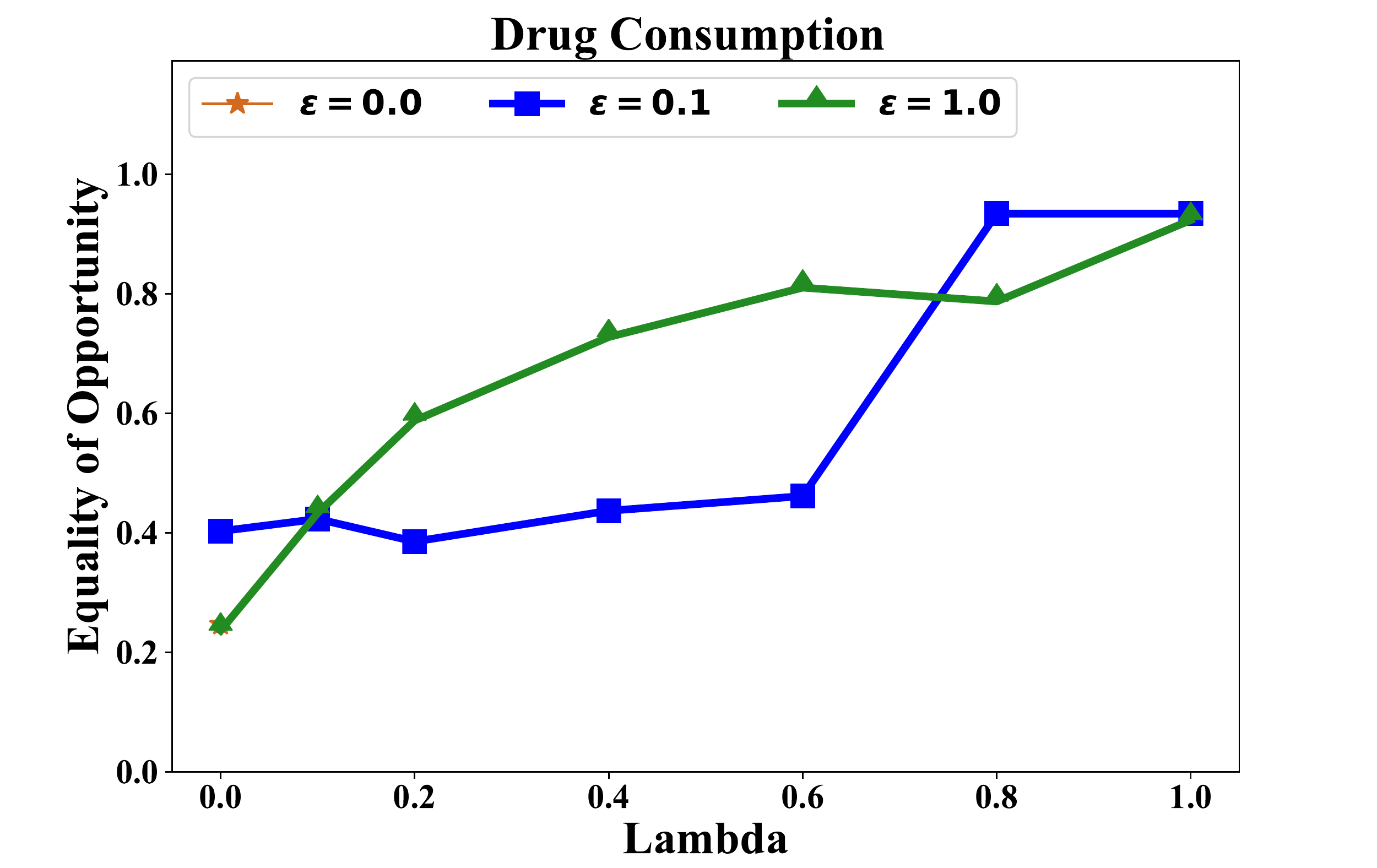}
           \caption{Results obtained for different lambda values for the IAF attack with regards to different fairness (SPD and EOD) and accuracy (test error) measures on three different datasets (German Credit, COMPAS, and Drug Consumption) with different $\epsilon$.} 
    \label{lambda_attack_results}
\end{figure*}
To evaluate our attacks, we compared them against an attack that does not consider fairness and only considers accuracy to show that such attacks are not necessarily effective for fairness, motivating the need for fairness attacks. Also, we compared our attacks in terms of how they attack accuracy as a measure versus  attacks that are specifically designed to target accuracy. We also compared our attacks to an attack that is optimized for fairness. 

The evaluated methods are listed below. In our experiments, the poisoned points are inversely proportional to class balance as also suggested in \cite{koh2018stronger}, so we made $(|\mathcal{D}_{c}^+|\epsilon)$ copies from the negative poison instance ($\mathcal{I}_{-}$) and $(|\mathcal{D}_{c}^-|\epsilon)$ copies from the positive poison instance ($\mathcal{I}_{+}$) in which $|\mathcal{D}_{c}^+|$ and $|\mathcal{D}_{c}^-|$ denote the number of positive and negative points in the clean data respectively. Hinge loss was used to control for accuracy for all the methods in our experiments as in \cite{koh2018stronger}. \\ 
\textbf{Influence Attack on Fairness (IAF)} In this paper, our influence attack on fairness is where the attack tries to maximize the covariance between the signed distance of feature vectors from the decision boundary to the sensitive features, which would then cause the attack to target and degrade fairness. In our experiments we set $\lambda =1$.

\textbf{Random Anchoring Attack (RAA)} The anchoring attack where a target point is picked at random. In this new set of attacks, the goal is to place poisoned points in the vicinity of the target points in which the poisoned and target points have the same demographic group but different labels. In our experiments we set $\tau=0$ indicating the closest vicinity.  

\textbf{Non-random Anchoring Attack (NRAA)} This attack builds upon the random anchoring attack; however, in this attack, the target point is not chosen randomly. In this attack, the point with the most neighbors similar to it (with the same demographic group and label) is chosen as the target point so that we can infect as many similar points to the target point as possible. This can be effective because we are infecting more targeted points; however, in some cases it might be less effective since more poisoned points may be needed in order to achieve the goal of infecting many points and shifting the decision boundary. In our experiments we set $\tau=0$. 

\textbf{Influence Attack (Koh et al.)} This is a type of attack that is targeted only toward affecting accuracy \cite{koh2018stronger,koh2017understanding}. The reason we include this type of attack along with attacks targeted toward fairness is that it can help us understand how attacks targeting only accuracy affect fairness measures. Attacks of this nature can also serve as a good comparison because they show the effect of attacks on accuracy; because this attack is specifically designed to target accuracy, it can be a strong method to compare against.

\textbf{Poisoning Attack Against Algorithmic Fairness (Solans et al.)} In \cite{solans2020poisoning}, the authors propose a loss function that claims to target fairness measures. We utilized the loss introduced in this paper as depicted below in equation \eqref{baseline_loss} in the influence attack from \cite{koh2018stronger,koh2017understanding} and compared it to our proposed attacks. The goal of \cite{solans2020poisoning} was to incorporate the loss in \eqref{baseline_loss} into an attack strategy that would maximize the loss; thus, we  incorporated this loss into the influence attack \cite{koh2018stronger,koh2017understanding}, which we found to be a  strong attack strategy in maximizing the loss and also the same attack strategy used in our influence attack on fairness. In our experiments, we utilized the same $\lambda$ value as proposed in \cite{solans2020poisoning} to balance the class priors.
\begin{align*}
 L_{adv}(\hat{\theta};\mathcal{D}_{test}) = &\underbrace{\sum_{k=1}^p \ell(\hat{\theta};x_k,y_k)}_\text{disadvantaged} + \lambda \underbrace{\sum_{j=1}^m \ell(\hat{\theta};x_j,y_j)}_\text{advantaged} \\
& where \;\; \lambda = \frac{p}{m}.
\numberthis \label{baseline_loss}
\end{align*}
\subsection{Results}
The results in Figure \ref{attack_results} demonstrate that the influence attack (Koh et al.), although performing remarkably well in attacking accuracy, does not attack fairness well. The results also confirm that our influence attack on fairness method outperforms (Solans et al.) \cite{solans2020poisoning} in affecting fairness measures, and anchoring attack outperforms (Solans et al.) \cite{solans2020poisoning} in affecting fairness measures in most of the cases. One can observe that influence attack on fairness is the most effective amongst all the attacks in attacking fairness measures.

Due to the nature of our influence attack on fairness loss function and its controlling parameter on accuracy and fairness, it can be utilized in scenarios where the adversary wants to maliciously harm the system in terms of accuracy, or fairness, or both. On the other hand, anchoring attacks can be utilized in places where the adversary wants to subtly harm accuracy with an effective harm on fairness. These types of attacks can be used by, e.g., adversaries who would want to gain profit off of biasing decisions for their benefit; thus, to remain less detectable they do not harm accuracy. Although it is possible that anchoring attack can harm accuracy to a higher degree, as shown empirically in our results, it is less likely that anchoring attack is able to degrade accuracy by a large amount in practice for real world datasets.

In addition, in Figure \ref{lambda_attack_results} we demonstrate the effect of our regularized loss in the influence attack on fairness. The results show that with the increase of lambda the attack affects fairness measures more as expected from the loss; however, for the lower lambda values the attack acts similar to the original influence attack targeted towards accuracy. The results also show that higher epsilon values highlight the behavior of the loss more as expected such that for high epsilon value of 1 the changes are more significant with modifications to the lambda value in the loss function, while less subtle for lower epsilon values such as 0.1. 
\section{Related Work}
Here, we cover related work from both fair machine learning as well as adversarial machine learning research.
\subsection{Adversarial Machine Learning}
Research in adversarial machine learning is mostly focused on designing defenses and attacks against machine learning models \cite{NIPS2017_6943,chakraborty2018adversarial,li2018security}. Ultimately, the goal is for  machine learning models to be robust toward malicious activities designed by adversaries. Thus, it is important to consider both sides of the spectrum in terms of designing the attacks and defenses that can overcome the attacks. In adversarial machine learning, different types of attacks, such as data poisoning and evasion attacks, exist. In evasion attacks, the goal is to come up with adversarial examples that are imperceptible to human eye but can deceive benign machine learning models during test time \cite{biggio2013evasion,moosavi2016deepfool,DBLP:journals/corr/GoodfellowSS14}. On the other hand, in data poisoning attacks, the goal is to manipulate the training data--via adding, removing, or changing instances--so that the learned model is malicious \cite{10.5555/3042573.3042761,shafahi2018poison}. Different algorithms and approaches have been proposed for poisoning attacks focusing on accuracy as the performance measure \cite{10.5555/3042573.3042761,shafahi2018poison}. In this paper, we also focused on data poisoning attack while considering fairness as a performance measure in addition to accuracy.
\subsection{Fair Machine Learning}
Research in fair machine learning has gained attention recently, with   many active research areas. For instance, some work introduces new definitions and measures for fairness \cite{dwork2012fairness,hardt2016equality,NIPS2017_6995,mehrabi2020statistical}. \cite{10.1145/3194770.3194776} has a complete list of the definitions on fairness. Other work utilizes these definitions and tries to design and learn fair classification \cite{zafar2015learning,pmlr-v97-ustun19a}, regression \cite{agarwal2019fair}, and representations \cite{moyer2018invariant}. The battle to mitigate unfairness can happen in different phases. Some target making the data more fair \cite{zhang2017achieving}, while others target the algorithms \cite{zafar2015learning}. These mitigation techniques can also vary in when and how they are  applied. For instance, some approaches are pre-processing techniques \cite{Kamiran2012} in which the focus is to remove discrimination from the data before the learning phase. Others try to impose fairness during training via incorporation of fair loss functions or other approaches during the training phase, known as in-processing \cite{kamishima2012fairness}, while some are post-processing approaches \cite{NIPS2017_7151} in which the model is treated as a black box system and discrimination removal is performed on the output of the model. \citet{mehrabi2019survey} performs a literature review of fair machine learning research in different subject domains, which can be referenced for more detail. In our work, we  utilize some of the definitions and measures widely used in fair machine learning research \cite{dwork2012fairness,hardt2016equality} in measuring the performance of our attacks with regard to fairness. We were also inspired by some loss functions introduced in fair classification tasks in one of our attacks \cite{zafar2015learning}.
\subsection{Adversarial Fair Machine Learning}
The rapid and significant growth of research in algorithmic fairness highlights the importance of machine learning models being fair and robust toward any unfair behavior. To this end, it is important to think about attacks that can make models unfair in order to strengthen models against such  attacks. This recent and interesting line of work  combines the two fields of fair and adversarial machine learning. The only work we are aware of that proposes poisoning attacks against algorithmic fairness is \cite{solans2020poisoning}. In \cite{solans2020poisoning}, the authors propose an attack that targets fairness. We compared this attack with our two newly proposed attacks using three real world datasets. Our anchoring attack does not rely on any loss function making it different in nature with the previous work. In our influence attack on fairness we introduce a new loss function different than the previous work which is more in line with fairness literature and work done in fairness domain making our attack more intuitive. In addition, our influence attack on fairness is able to control a fairness-accuracy trade-off with the hyper-parameter involved in its loss function which is also shown in Figure \ref{lambda_attack_results} as an additional experimental result. This line of work can bring researchers from both fields closer and inspire new and interesting research problems. Another interdisciplinary research field combining concepts from fairness and privacy includes the differential privacy line of work \cite{dwork2008differential,bagdasaryan2019differential,pmlr-v97-jagielski19a,pujol2020fair}.

\section{Conclusion and Future Work}
In this work, we introduced two families of poisoning attacks that can target fairness. We  showed the effectiveness of these attacks through experimentation on different real world datasets with different measures. Our influence attack on fairness (IAF) used the attack strategy as in influence attack \cite{koh2018stronger,koh2017understanding}. As an extension, we modified the loss function so that it can harm fairness as well as accuracy. Furthermore, we explore an attack strategy called the ``anchoring attack'' that harms fairness by placing poisoned points near target points in order to bias the outcome. Our paper also introduced two ways of sampling these target points. A direct extension of this approach is to explore other methods of sampling points to increase the effectiveness of this attack. The introduced attacks each have their own advantages and disadvantages. The goal was to design attacks that can complement each other. For instance, influence attack on fairness which is gradient-based can be slow. Anchoring attack, however, does not use gradients and is considerably faster. Further, while influence attack on fairness targets fairness harshly, anchoring attack is more subtle. And if anchoring attack can not explicitly control for accuracy fairness trade-off, influence attack on fairness can control this trade-off. 

This work points out several important angles for future research. Some important extensions are as follows: what other ways can machine learning systems be harmed by data poisoning attacks? How can we design and adapt defenses that can be effective against malicious attacks targeting fairness? Another question worth pursuing is from the perspective of the defender. Can current defenses against accuracy attacks be useful against the types of attacks that target fairness. If not, how do we adapt defenders so that they can prevent fairness attacks? These questions can help us design more fair, and accurate models that are robust to poisoning attacks. By extension, one can also think about stronger attacks against fairness than ours. We anticipate that continuing to blend the fields of adversarial and fair machine learning can create interdisciplinary ideas that can help us develop more robust and fair machine learning models. 

\section{Acknowledgments}
This material is based upon work supported by the Defense Advanced Research Projects Agency (DARPA) under Agreement No. HR0011890019. We thank the anonymous reviewers and Mozhdeh Gheini for their constructive feedback.
\section{Ethics Statement}
This paper furthers ethics in the machine learning community in two major ways:
\begin{itemize}
    \item 
    Despite extensive research in adversarial machine learning,  not much attention has been given to scenarios where fairness is a possible target of deliberate attacks. We suggest that fairness metrics are as important as accuracy, because they can be manipulated in sensitive environments to achieve malicious goals. Our work points out potential vulnerabilities of  machine learning models against fairness-targeting attacks. This line of research can raise awareness, and motivate researchers to introduce methods to mitigate harmful effects of adversarial attacks on fairness. The attacks proposed in this paper are meant to ensure the robustness of fairness in machine learning applications. Nevertheless, we acknowledge that in the wrong hands these type of tools could enable an attacker to harm fairness in extant machine learning systems. 
    \item Fairness and adversarial machine learning are both very important research areas with major safety, security, and ethics implications both within and beyond on the AI/machine learning community. This work combines ideas from both adversarial and fair machine learning, and will hopefully facilitate collaboration among researchers from both communities, eventually leading to more robust and fair machine learning models.
\end{itemize}

\bibstyle{aaai21}
\bibliography{Ref}
\end{document}